%% file: main.tex
\documentclass[10pt,twocolumn,letterpaper]{article}

\usepackage[pagenumbers]{cvpr} 


\definecolor{cvprblue}{rgb}{0.21,0.49,0.74}
\definecolor{firstred}{HTML}{FFB3B3}
\definecolor{secondorange}{HTML}{FFD9B3}
\definecolor{thirdyellow}{HTML}{FFFFB3}
\usepackage[pagebackref,breaklinks,colorlinks,allcolors=cvprblue]{hyperref}

\usepackage[capitalize]{cleveref}
\crefname{section}{Sec.}{Secs.}
\Crefname{section}{Section}{Sections}
\Crefname{table}{Table}{Tables}
\crefname{table}{Tab.}{Tabs.}

\usepackage[utf8]{inputenc} 
\usepackage[T1]{fontenc}    
\usepackage{hyperref}       
\usepackage{url}            
\usepackage{booktabs}       
\usepackage{amsfonts}       
\usepackage{nicefrac}       
\usepackage{microtype}      
\usepackage{xcolor}         
\usepackage{amsmath}
\usepackage{bm}
\usepackage{pifont}         

\usepackage{multirow}
\usepackage{makecell}
\usepackage{graphicx}
\usepackage{subcaption}
\usepackage{booktabs}
\usepackage{tikz}
\usepackage{bbding}
\usepackage{overpic}
\usepackage{rotating}
\usepackage{colortbl} 

\usepackage{float}
\usepackage{wrapfig}
\usepackage{threeparttable}
\usepackage[numbers]{natbib}

\graphicspath{ {figs/} }

\newcommand{\Loss}{\mathcal{L}}

\newcommand{\figref}[1]{Fig.~\ref{#1}}

\newcommand{\tabref}[1]{Tab.~\ref{#1}}
\newcommand{\equref}[1]{Eq.~\ref{#1}}
\newcommand{\apref}[1]{\ref*{#1}}
\newcommand{\nolinkfigref}[1]{%
    \begingroup
    \hypersetup{hidelinks}%
    Fig.~\ref{#1}%
    \endgroup
}
\newcommand{\nolinktabref}[1]{%
    \begingroup
    \hypersetup{hidelinks}%
    Tab.~\ref{#1}%
    \endgroup
}

\newcommand{\sdf}{\mathsf{S}_\theta}
\newcommand{\netdiff}{\mathsf{M}_d}
\newcommand{\netspec}{\mathsf{M}_s}
\newcommand{\netcolor}{\mathsf{M}_c}
\newcommand{\netspecalb}{\mathsf{M}_{sa}}
\newcommand{\netvis}{\mathsf{M}_\vis}
\newcommand{\netind}{\mathsf{M}_\text{ind}}
\newcommand{\modelname}{Factored-NeuS}

\newcommand{\normal}{\bm{n}}
\newcommand{\R}{\mathbb{R}}
\newcommand{\Ss}{\mathbb{S}}
\newcommand{\dir}{\boldsymbol{\omega}} 
\newcommand{\diri}{\dir_i} 
\newcommand{\diro}{\dir_o} 
\newcommand{\dirr}{\dir_r} 
\newcommand{\feat}{\bm{v}_f} 
\newcommand{\weight}{w} 
\newcommand{\coord}{\bm{x}} 
\newcommand{\coordset}{\mathcal{X}}
\newcommand{\ray}{\bm{r}}
\newcommand{\rayset}{\mathcal{R}}
\newcommand{\origin}{\bm{o}}
\newcommand{\raydir}{\bm{d}}
\newcommand{\vis}{\nu} 
\newcommand{\diffalb}{\bm{d}_{a}} 
\newcommand{\specalb}{\bm{s}_{a}} 
\newcommand{\specbrdf}{f^a_s} 
\newcommand{\colord}{\bm{c}_d}
\newcommand{\colors}{\bm{c}_s}
\newcommand{\colorsur}{{C}^\text{sur}}
\newcommand{\colorvol}{{C}^\text{vol}}
\newcommand{\colorgt}{{C}^\text{gt}}
\newcommand{\izero}{i'}
\newcommand{\latent}{\bm{z}}
\newcommand{\sgaxisind}{\bm{\xi}^i}
\newcommand{\sgsharpind}{\lambda^i}
\newcommand{\sgamplind}{\bm{\mu}^i}
\newcommand{\sgaxisenv}{\bm{\xi}^e}
\newcommand{\sgsharpenv}{\lambda^e}
\newcommand{\sgamplenv}{\bm{\mu}^e}

\newcommand{\Lvol}{\Loss_\text{vol}}
\newcommand{\Lsur}{\Loss_\text{sur}}
\newcommand{\Lreg}{\Loss_\text{reg}}

\newcommand{\Lsurcoef}{\lambda_\text{sur}}
\newcommand{\Lregcoef}{\lambda_\text{reg}}

\def\paperID{13280} 
\def\confName{CVPR}
\def\confYear{2025}

\title{\modelname: Reconstructing Surfaces, Illumination, and Materials of Possibly Glossy Objects}

\author{
Yue Fan\textsuperscript{1}
\qquad
Ningjing Fan\textsuperscript{1}
\qquad
Ivan Skorokhodov\textsuperscript{2}
\qquad
Oleg Voynov\textsuperscript{3,}\textsuperscript{4}
\qquad
Savva Ignatyev\textsuperscript{3}
\\
Evgeny Burnaev\textsuperscript{3,}\textsuperscript{4}
\qquad
Peter Wonka\textsuperscript{2}
\qquad Yiqun Wang\textsuperscript{1}\thanks{Corresponding author: yiqun.wang@cqu.edu.cn}
\\ 
\small {
\textsuperscript{1}Chongqing University
\qquad
\textsuperscript{2}KAUST
\qquad
\textsuperscript{3}Skoltech
\qquad
\textsuperscript{4}AIRI
}
}

\begin{document}
\maketitle

\begin{abstract}
We develop a method that recovers the surface, materials, and illumination of a scene from its posed multi-view images.
In contrast to prior work, it does not require any additional data and can handle glossy objects or bright lighting.
It is a progressive inverse rendering approach, which consists of three stages.
In the first stage, we reconstruct the scene radiance and signed distance function (SDF) with a novel regularization strategy for specular reflections.
We propose to explain a pixel color using both surface and volume rendering jointly, which allows for handling complex view-dependent lighting effects for surface reconstruction.
In the second stage, we distill light visibility and indirect illumination from the learned SDF and radiance field using learnable mapping functions.
Finally, we design a method for estimating the ratio of incoming direct light reflected in a specular manner and use it to reconstruct the materials and direct illumination.
Experimental results demonstrate that the proposed method outperforms the current state-of-the-art in recovering surfaces, materials, and lighting without relying on any additional data.
\end{abstract}

\section{Introduction}

Reconstructing shape, material, and lighting from multiple views has wide applications in computer vision, virtual reality, augmented reality, and shape analysis.
The emergence of neural radiance fields~\citep{Nerf} provides a framework for high-quality scene reconstruction. Subsequently, many works~\citep{Unisurf,NeuS,Volsdf,HF-NeuS,geo-neus} have incorporated implicit neural surfaces into neural radiance fields, further enhancing the quality of surface reconstruction from multi-views. Recently, several works~\citep{NVDiffrec,Nerfactor,Physg,sun2023neural,Indirect} have utilized coordinate-based networks to predict materials and learned parameters to represent illumination, followed by synthesizing image color using physically-based rendering equations to achieve material and lighting reconstruction. However, these methods typically do not fully consider the interdependence between different components, leading to the following issues with glossy surfaces when using real data. 

\begin{figure*}[t]
    \centering
    \vspace{-0.5cm}
    \begin{overpic}[width=\linewidth]{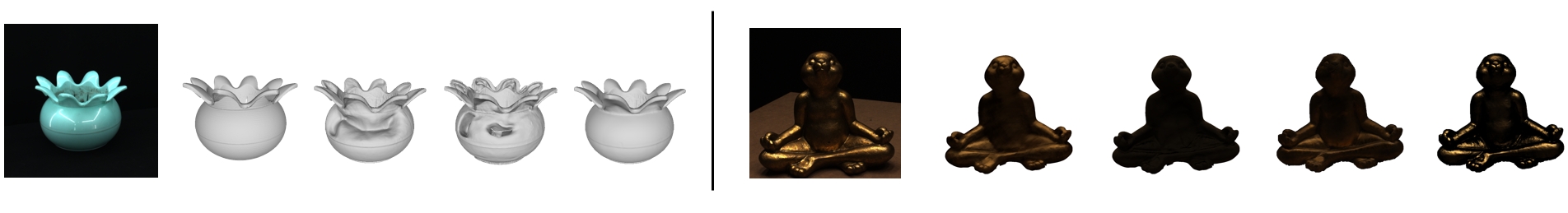}
    \put(3.7,0){\fontsize{7pt}{0}\selectfont\color{black}{Image}}
    \put(12.7,0){\fontsize{7pt}{0}\selectfont\color{black}{GT Mesh}}
    \put(22.5,0){\fontsize{7pt}{0}\selectfont\color{black}{NeuS}}
    \put(29,0){\fontsize{7pt}{0}\selectfont\color{black}{Geo-NeuS}}
    \put(39,0){\fontsize{7pt}{0}\selectfont\color{black}{Ours}}

    \put(51,0){\fontsize{7pt}{0}\selectfont\color{black}{Image}}
    \put(62.5,0){\fontsize{7pt}{0}\selectfont\color{black}{IndiSG}}
    \put(62.5,-1.5){\fontsize{7pt}{0}\selectfont\color{black}{Diffuse}}
    \put(72.7,0){\fontsize{7pt}{0}\selectfont\color{black}{IndiSG}}
    \put(72,-1.5){\fontsize{7pt}{0}\selectfont\color{black}{Specular}}
    \put(84,0){\fontsize{7pt}{0}\selectfont\color{black}{Ours}}
    \put(83,-1.5){\fontsize{7pt}{0}\selectfont\color{black}{Diffuse}}
    \put(94.5, 0){\fontsize{7pt}{0}\selectfont\color{black}{Ours}}
    \put(93.5,-1.5){\fontsize{7pt}{0}\selectfont\color{black}{Specular}}
    \end{overpic}
    \vspace{-0.2cm}
    \caption{Left: Geometry visualization for NeuS, Geo-NeuS and our method on the Pot scene from SK3D dataset. Existing surface reconstruction methods struggle to recover the correct geometry of glossy objects due to the complex view-dependent effects they induce. The weak color model of these methods compels to represent such effects through concave geometric deformations rather than proper view-dependent radiance, leading to shape artifacts. In contrast, our method can correctly reconstruct a highly reflective surface due to our joint appearance, diffuse, and specular color training strategy. Right: Visualization of the recovered diffuse color component on the Bunny scene from DTU for IndiSG~\citep{Indirect} and our method. Existing inverse rendering methods overestimate the diffuse material component in the presence of specular highlights. Our regularization strategy allows us to properly disentangle the color into diffuse and specular components.}
    \label{fig:motivation}
    \vspace{-0.3cm}   
\end{figure*}

First, surfaces with glossy materials typically result in highlights. The best current methods~\citep{NeuS,HF-NeuS,geo-neus} for reconstructing implicit neural surfaces rarely consider material information and directly reconstruct surfaces. The surface parameters can then be frozen for subsequent material reconstruction. Since neural radiance fields typically model such inconsistent colors as bumpy surfaces as shown in \figref{fig:motivation} left, the artifacts from surface reconstruction will affect material reconstruction if surfaces and materials are reconstructed sequentially. Second, a glossy surface can affect the decomposition of the reflected radiance into a diffuse component and a specular component. Typically, the specular component leaks into the diffuse component, resulting in inaccurate modeling as shown in \figref{fig:motivation} right. Third, focusing on synthetic data makes it easier to incorporate complex physically-based rendering algorithms~\citep{wardbrdf,Disney}, but they may not be robust enough to work on real data.

In this work, we consider the impact of glossy surfaces on surface and material reconstruction based on implicit representation. Although existing methods based on explicit representations, such as 3D Gaussian Splatting (3DGS)~\citep{kerbl20233d}, have made progress in surface reconstruction~\citep{Huang2DGS2024,Yu2024GOF,chen2024pgsr} and inverse rendering~\citep{R3DG2023,jiang2024gaussianshader,shi2023gir,liang2024gs,ye2024gsdr}, the unique advantages of implicit representations, such as low storage requirements and smooth surface modeling, still make them worthy of being explored as a separate research branch. This paper focuses solely on methods based on implicit representation and further extends research in reconstructing glossy objects.
To better handle glossy surfaces, we jointly use surface and volume rendering. Volume rendering does not decompose the reflected radiance, while surface rendering considers the diffuse and specular radiance separately. 
The learned SDF network enables accurate radiance reconstruction in volume rendering while also being capable of providing correct surface information in surface rendering to model specular and diffuse components into directional and non-directional parts. Both work together to regularize accurate and efficient SDF surface reconstruction of glossy objects. 
To better recover diffuse and specular components, we estimate the ratio of incoming light reflected in a specular manner.
By introducing this parameter into a Spherical Gaussian representation of the BRDF for inverse rendering, we can better model the reflection of glossy surfaces and decompose more accurate diffuse albedo information.
Furthermore, we propose predicting continuous light visibility for signed distance functions to further enhance the quality of reconstructed materials and illumination.
Our experimental results have shown that our factorization of surface, materials, and illumination achieves state-of-the-art performance on both synthetic and real datasets.

We believe our approach outperforms recent work because it was developed primarily on real data. 
The fundamental challenge in material and lighting reconstruction is the lack of available ground truth information for real datasets. 
Our solution was to work with real data and improve surface reconstruction as our main metric by experimenting with different materials and lighting decompositions as a regularizer, indirectly measuring the success of material and lighting reconstruction through surface metrics.
By contrast, most recent works~\citep{dip,Physg,Indirect} use surface reconstruction and real data more as an afterthought, focusing instead on increasingly complex material and lighting reconstruction with synthetic data. However, we believe that this typically does not translate as well to real data as our approach.

\section{Related work}
\label{related}

\noindent\textbf{Neural radiance fields}.
NeRF~\citep{Nerf} is a seminal work in 3D reconstruction. 
Important improvements were proposed by Mip-NeRF~\citep{Mip} and Mip-NeRF360~\citep{mipnerf360}. One line of work explores the combination of different data structures with MLPs, such as factored volumes ~\citep{EG3D,Tensorf,petneus} or voxels~\citep{Instant-ngp,Kilonerf,Plenoxels}.
Multiple approaches take a step towards extending neural radiance fields to reconstruct material information~\citep{Nerfren,Refnerf,Refneus,s3nerf,Bakedsdf}.

\noindent\textbf{Implicit neural surfaces}.
Implicit neural surfaces are typically represented by occupancy functions or signed distance fields (SDFs).
Some early works~\citep{IM-Net, Occupancy_net, Deepsdf} take point clouds as input and output implicit neural surface representations. 
Many works have studied how to obtain implicit neural surfaces from images, initially focusing on surface rendering only~\citep{DVR, IDR}. Subsequent methods followed NeRF to employ volume rendering, e.g. UNISURF~\citep{Unisurf}, VolSDF~\citep{Volsdf}, NeuS~\citep{NeuS}, HF-NeuS~\citep{HF-NeuS}, and Geo-NeuS~\citep{geo-neus}.

\noindent\textbf{Joint reconstruction of surface, material, and illumination}.
Ideally, we would like to jointly reconstruct the 3D geometry, material properties, and lighting conditions of a scene from 2D images. 
Several methods employ strategies to simplify the problem such as assuming known lighting conditions (NeRV~\citep{Nerv} and NeRD~\citep{Nerd}) or pre-training (ENVIDR~\citep{Envidr}). 
PhySG~\citep{Physg}, NeRFactor~\citep{Nerfactor}, and NeROIC~\citep{Neroic} use Spherical Gaussians, point light sources, and spherical harmonics, respectively, to decompose unknown lighting from a set of images.
Using an illumination integration network, Neural-PIL~\citep {Neuralpil} further reduces the computational cost of lighting integration. IRON~\citep{Iron} uses SDF-based volume rendering methods to obtain better geometric details in the shape recovery stage. NVDiffrec~\citep{NVDiffrec} explicitly extracts triangle mesh from tetrahedral representation for better material and lighting modeling.
IndiSG~\citep{Indirect} uses Spherical Gaussians to represent indirect illumination and achieves good lighting decomposition results.
Some works~\citep{TensoIR,Nefii,NeILF++,NeMF,dip} continue to improve the efficiency and quality of inverse rendering but do not consider cases with a glossy appearance.
NeAI~\citep{NeAI} proposes neural ambient illumination to enhance the rendering quality of glossy appearance. NeRO~\citep{nero} highlights its good performance in reconstructing reflective objects.
Ref-NeuS~\cite{Refneus} learns specular reflection intensity and attenuates the impact of pixels affected by specular reflection.
Despite a lot of recent activity in this area, existing frameworks still struggle to effectively reconstruct reflective or glossy surfaces, lighting, and material information directly from images, especially real-world captured images.

\begin{figure*}[t]
  \centering
  \setlength{\abovecaptionskip}{0.cm}
  \vspace{-0.5cm}   
  \includegraphics[width=\linewidth]{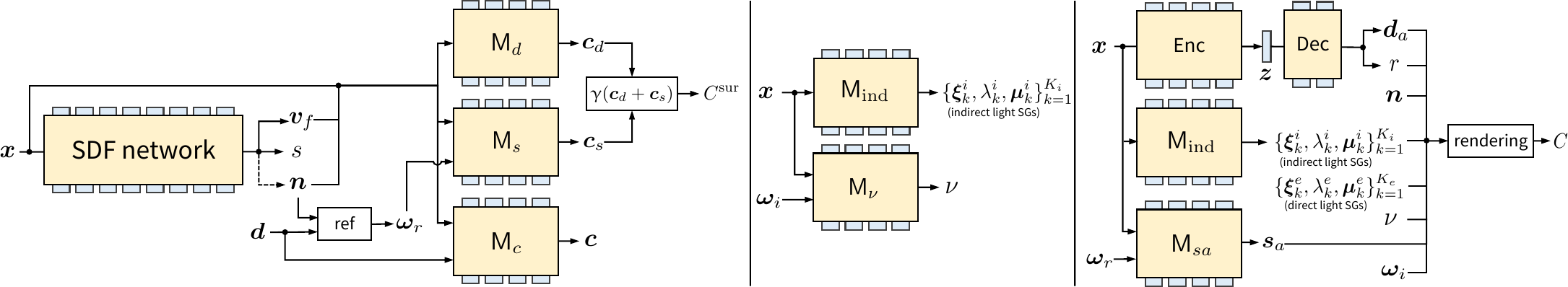}
  \vspace{-0.2cm}
  \caption{Overview for Stage 1 (left), Stage 2 (mid), and Stage 3 (right) training pipelines. The first stage (left) trains the SDF network $\sdf$ which outputs a feature vector $\feat \in \R^{256}$, SDF value $s \in \R$, and normal $\normal \in \Ss^2$ (as a normalized gradient of $s$; denoted via the dashed line); diffuse and specular surface color networks $\netdiff$ and $\netspec$ produce their respective colors $\colord, \colors \in \R^3$ via surface rendering, which are then combined through tone mapping $\gamma(\cdot)$ to get the final surface color $\colorsur \in \R^3$; volumetric color network $\netcolor$ produces the volumetrically rendered color $\colorvol \in \R^3$. The $\mathsf{ref}$ operation denotes computation of the reflection direction $\dirr \in \Ss^2$ from normal $\normal$ and ray direction $\dir \in \Ss^2$. In the second stage (mid), indirect light network $\netind$ and light visibility network $\netvis$ produce their respective indirect light SGs and light visibility $\vis$. In the third stage (right), we optimize the material BRDF auto-encoder with the sparsity constraint~\citep{Indirect}, our novel specular albedo network $\netspecalb$, and the indirect illumination network $\netind$. See Sec~\ref{method} for details. }
  \label{fig:stage1_pipeline}
  \vspace{-0.4cm}
\end{figure*}

\section{Method}
\label{method}

Our framework has three training stages to gradually decompose the shape, materials, and illumination. The input to our framework is a set of images.
In the first stage, we reconstruct the surface from a (possibly glossy) appearance by jointly using volume rendering and surface rendering.
After that, we use the reconstructed radiance field to extract direct illumination visibility and indirect illumination.
In the final stage, we recover the direct illumination map and materials' bidirectional reflectance distribution function (BRDF), proposing to use a BRDF with a learnable specular albedo.

\subsection{Stage 1: Surface reconstruction from glossy appearance}
\label{sec:surface}

Current inverse rendering methods first recover implicit neural surfaces, typically represented as SDFs, from multi-view images to recover shape information, then freeze the parameters of neural surfaces to further recover the material.
However, this approach does not consider specular reflections that produce highlights and often models this inconsistent color as bumpy surface geometry as depicted in \figref{fig:motivation} left.
This incorrect surface reconstruction has a negative impact on subsequent material reconstruction.
We propose a neural surface reconstruction method that considers the appearance, diffuse color, and specular color of glossy surfaces at the same time, whose architecture is given in \figref{fig:stage1_pipeline}.
Our inspiration comes from the following observations. First, according to Geo-NeuS~\citep{geo-neus}, using SDF point cloud supervision can make the colors of surface points and volume rendering more similar. We abandoned the idea of using additional surface points to supervise SDFs and directly used two different MLPs to predict the surface rendering and volume rendering results and narrow the gap between these two colors using network training. In addition, when modeling glossy surfaces, Ref-NeRF~\cite{Refnerf} proposes a method of decomposing reflected radiance into diffuse and specular components, which can better model the glossy appearance. However, this approach is unstable when directly applied to reconstruct implicit surfaces. We propose to simultaneously optimize the radiance from the volumetric rendering and the surface rendering. For surface rendering, we further split the reflected radiance into a diffuse and a specular component. This can achieve an improved surface reconstruction of glossy surfaces.

\noindent\textbf{Shape representation}.
We model shape as a signed distance function $\sdf: \coord \mapsto (s,\feat)$, which maps a 3D point $\coord \in \R^3$ to its signed distance value $s \in \R$ and a feature vector $\feat \in \R^{256}$.
SDF allows computing a normal $\normal$ directly by calculating the gradient: $\normal = \nabla \sdf(\coord) / \|\nabla \sdf(\coord) \|$.

\noindent\textbf{Synthesize appearance.}  Learning implicit neural surfaces from multi-view images often requires synthesizing appearance/color to optimize the underlying surface. The recent use of volume rendering in NeuS~\citep{NeuS} has been shown to better reconstruct surfaces. According to \equref{eq:rendering_equ} in Appx~\apref{ap:preliminaries}, the discretization formula for volume rendering is $\colorvol=\sum_{i=1}^nT_i\alpha_i\bm{c_i} = \sum_{i=1}^n\weight_i\bm{c_i}$ with $n$ sampled points $\{\ray(t_i)\}_{i=1}^n$ on the ray.
where $\alpha_i=\max(\Phi_s(\sdf(\ray(t_i)))- \Phi_s(\sdf(\ray(t_{i+1})))/{\Phi_s(\sdf(\ray(t_i)))},0)$, which is discrete opacity values following NeuS, where $\Phi_s = 1/(1 + e^{-x})$ is a sigmoid function and $T_i=\prod_{j=1}^{i-1}(1-\alpha_j)$ is the discrete transparency.
Similar to the continuous case, we can also define discrete weights $\weight_i=T_i\alpha_i$.
To compute color $\bm{c_i}$ on the point $\ray(t_i)$, we define a color mapping $\netcolor: ({\coord,\normal, \raydir},\feat) \mapsto \bm{c}$ from any 3D point $\coord$ given its feature vector $\feat$, normal $\normal$ and ray direction $\raydir$. 

\noindent\textbf{Synthesize diffuse and specular components.}
We also synthesize separated diffuse and specular components as proposed in~\citep{hedman2021baking,Bakedsdf,Envidr}, but uniquely, we simultaneously preserve the synthesis of the overall appearance.
This idea comes from surface rendering, which better handles surface reflections.
From \equref{eq:surface_rendering} in Appx~\apref{ap:preliminaries}, the radiance $L_o$ of surface point $\coord$ and outgoing viewing direction $\diro$ can be decomposed into two parts: diffuse and specular radiance.
\begin{align}
L_{o}(\coord, \diro)&=\frac{\diffalb}{\pi} \int_{\Omega} L_{i}(\coord, \diri)(\diri \cdot \normal) d \diri \\
&+\int_{\Omega} f_{s}(\coord, \diri, \diro ) L_{i}(\coord, \diri)(\diri \cdot \normal) d\diri \label{eq:sep_render} \\
&= \netdiff(\coord,\normal) + \netspec(\coord,\diro,\normal)
\label{eq:separation}
\end{align}
We define two neural networks to predict diffuse and specular components separately.
We use the term diffuse radiance to refer to the component of the reflected radiance that stems from a diffuse surface reflection.
We define a mapping $\netdiff: ({\coord,\normal},\feat)\mapsto \colord$ for diffuse radiance that maps surface points $\coord$, surface normals $\normal$, and feature vectors $\feat$ to diffuse radiance. 
For simplicity, we assume that the diffuse radiance is not related to the outgoing viewing direction $\diro$.

We use the term specular radiance to describe the non-diffuse (view-direction dependent) component of the reflected radiance.
Ref-NeRF~\citep{Refnerf} proposes to model the glossy appearance using the reflection direction instead of the viewing one.
However, from \equref{eq:separation}, we can observe that specular radiance is also highly dependent on the surface normal, which is particularly important when reconstructing SDF.
In contrast to Ref-NeRF, we further condition specular radiance on the surface normal.
Therefore, we define specular radiance $\netspec: ({\coord,\dirr,\normal},\feat)\mapsto \colors$, which maps surface points $\coord$, reflection direction $\dirr$, surface normals $\normal$, and feature vectors $\feat$ to specular radiance, where $\dirr=2(\diro \cdot \normal) \normal-\diro$.

Surface rendering focuses the rendering process on the surface, allowing for a better understanding of highlights on the surface compared to volume rendering, but requires calculating surface points. 
We sample $n$ points on the ray $\{\ray(t_i) | i=1,...,n\}$. We query the sampled points to find the first point $\ray(t_\izero)$ whose SDF value is less than zero $\sdf(\ray(t_\izero))<0$.
Then the point $\ray(t_{\izero-1})$ sampled before $\ray(t_\izero)$ has the SDF value greater than or equal to zero $\sdf(\ray(t_{\izero-1}))\geq0$.
To account for the possibility of rays interacting with objects and having multiple intersection points, we select the first point with a negative SDF value to solve this issue.

We use two neural networks to predict the diffuse radiance and specular radiance of two sampling points $\ray(t_{\izero-1})$ and $ \ray(t_\izero)$. 
The diffuse radiance of the two points calculated by the diffuse network $\netdiff$ will be $\colord^{\izero-1}$ and $\colord^{\izero}$.
The specular radiance of the two points calculated by the specular network $\netspec$ will be $\colors^{\izero-1}$ and $\colors^{\izero}$.
Therefore, the diffuse radiance and specular radiance of the surface point $\coord$ can be calculated as follows.
\begin{align}
\colord = \netdiff(\coord,\normal) &= \frac{\weight_{\izero-1}\colord^{\izero-1}+\weight_\izero\colord^{\izero}}{\weight_{\izero-1}+\weight_\izero} \\
\colors = \netspec(\coord,\diro,\normal) &= \frac{\weight_{\izero-1}\colors^{\izero-1}+\weight_\izero\colors^{\izero}}{\weight_{\izero-1}+\weight_\izero}
\end{align}
The final radiance of the intersection of the ray and the surface is calculated by a tone mapping:
\begin{equation}
    \colorsur=\gamma(\colord+\colors)
\end{equation}
where $\gamma$ is a pre-defined tone mapping function that converts linear color to
sRGB~\citep{Refnerf} while ensuring that the resulting color values are within the valid range of [0, 1].

\noindent\textbf{Training strategies.} In our training process, we define three loss functions, namely volume radiance loss $\Lvol$, surface radiance loss $\Lsur$, and regularization loss $\Lreg$. 
The volume radiance loss $\Lvol$ is measured by calculating the $\Loss_1$ distance between the ground truth colors $\colorgt$ and the volume radiances $\colorvol$ of a subset of rays $\rayset$.
The surface radiance loss $\Lsur$ is measured by calculating the $\Loss_1$ distance between the ground truth colors $\colorgt$ and the surface radiances $\colorsur$.
$\Lreg$ is an Eikonal loss term on the sampled points.
We use weights $\Lsurcoef$ and $\Lregcoef$ to balance the impact of these three losses.
The overall training loss is as follows. See Appx~\apref{ap:training_strategies} for details of training strategies.
\begin{equation}
    \Loss = \Lvol + \Lsurcoef\Lsur + \Lregcoef\Lreg
\end{equation}

\subsection{Stage 2: Learning direct lighting visibility and indirect illumination}
\label{sec:visibility}
At this stage, we focus on predicting the lighting visibility and indirect illumination of a surface point $\coord$ under different incoming light direction $\diri$ using the SDF in the first stage. 

Visibility is an important factor in shadow computation.
It calculates the visibility of the current surface point $\coord$ in the direction of the incoming light $\diri$.
Path tracing of the SDF is commonly used to obtain a binary visibility (0 or 1) as used in IndiSG~\citep{Indirect}, but this kind of visibility is not friendly to network learning.
Inspired by NeRFactor~\citep{Nerfactor}, we propose to use an integral representation with the continuous weight function $\weight(t)$ (with a range from 0 to 1 and sourced from stage 1) for the SDF to express light visibility.
Specifically, we establish a neural network $\netvis: (\coord, {\diri}) \mapsto \vis$, that maps the surface point $\coord$ and incoming light direction $\diri$ to visibility, and the ground truth value of light visibility is obtained by integrating the weights $\weight_i$ of the SDF of sampling points along the incoming light direction and can be expressed as $\vis^{gt} = 1 - \sum_{i=1}^n\weight_i$. 

Indirect illumination refers to the light that is reflected or emitted from surfaces in a scene and then illuminates other surfaces, rather than directly coming from a light source, which contributes to the realism of rendered images.
Following IndiSG~\citep{Indirect}, we parameterize indirect illumination $I(\coord,{\diri})$ via $K_i=24$ Spherical Gaussians (SGs). For more details, see Appx~\apref{ap:details_s2}.

\subsection{Stage 3: Recovering materials and direct illumination}
\label{sec:materials}

Reconstructing good materials and lighting from scenes with highlights is a challenging task.
Following prior works \cite{Indirect, Physg}, we use the Ward BRDF model~\citep{wardbrdf} and represent BRDF $f_s(\diri \ |\  \boldsymbol{\xi}_s, \lambda_s, \boldsymbol{\mu}_s)$ via Spherical Gaussians~\citep{Physg}.
Direct (environment) illumination is represented using $K_e=128$ SGs:
\begin{equation}
E(\coord,\diri)=\sum_{k=1}^{K_e} E_k (\diri \ |\ \sgaxisenv_k, \sgsharpenv_k, \sgamplenv_k)
\end{equation}
and render diffuse radiance and specular radiance of direct illumination in a way similar to \equref{eq:sep_render}.
\begin{align}
L_{d}(\coord)&= \frac{\diffalb}{\pi} \sum_{k=1}^{K_e}(\vis(\coord, \diri) \otimes E_{k}(\diri)) \cdot (\diri \cdot \normal) \\
L_{s}(\coord, \diro)&=\sum_{k=1}^{K_e}(\specbrdf  \otimes \vis(\coord, \diri) \otimes E_{k}(\diri)) \cdot (\diri \cdot \normal)
\label{eq:SGs_separation}
\end{align}
where $\diffalb$ is diffuse albedo.

To reconstruct a more accurate specular reflection effect, we use an additional neural network $\netspecalb: (\coord,\dirr) \mapsto \specalb \in [0, 1]$ to predict the specular albedo.
The \textbf{modified} BRDF $\specbrdf$ is as follows:
\begin{equation}
\label{eq:spec_improvement}
    \specbrdf = \specalb \otimes f_s(\diri ; \boldsymbol{\xi}, \lambda, \boldsymbol{\mu}) = f_s(\diri ; \boldsymbol{\xi}, \lambda, \specalb \boldsymbol{\mu}) 
\end{equation}
For indirect illumination,  the radiance is extracted directly from another surface and does not consider light visibility. The diffuse radiance and specular radiance of indirect illumination are as follows
\begin{equation}
L^{\text{ind}}_{d}(\coord)=\frac{{\diffalb}}{\pi} \sum_{k=1}^{T} I_k (\coord,\diri) \cdot (\diri \cdot \normal)
\end{equation}
\begin{equation}
L^{\text{ind}}_{s}(\coord, \diro)=\sum_{k=1}^{T}( \specalb\otimes f_{s} ) \otimes I_k(\coord,\diri) \cdot (\diri \cdot \normal)
\end{equation}
Our final synthesized appearance is $C =L_d+L_s+L^{\text{ind}}_d+L^{\text{ind}}_s$ and supervised via an $\Loss_1$ RGB loss.

\section{Experiments}\label{sec:experiments}

\subsection{Evaluation setup}

\noindent\textbf{Datasets.}
To evaluate the quality of surface reconstruction, we use the SK3D~\citep{SK3D}, DTU~\citep{DTU}, Shiny~\citep{Refnerf}, and Glossy~\citep{nero} datasets.
SK3D and DTU are two real-world captured datasets, while Shiny is synthetic.
In the SK3D dataset, 100 views are provided for each scene.
Compared to DTU, SK3D contains more reflective objects with complex view-dependent lighting effects.
From SK3D, we select 4 glossy surface scenes with high levels of glare.
For DTU, we select 4 scenes with high specularities to evaluate our method in terms of surface quality quantitatively and material decomposition qualitatively. 
The Shiny dataset has 5 different glossy objects rendered in Blender under conditions similar to the NeRF-synthetic dataset~\cite{Nerf}. 
Glossy~\citep{nero} dataset contains both real and synthetic scenes.

To evaluate the effectiveness of material and lighting reconstruction, we use the dataset provided by IndiSG~\citep{Indirect}, which has self-occlusion and complex materials.
To evaluate the quality of material decomposition, the dataset also provides diffuse albedo, roughness, and masks for testing.


\noindent\textbf{Baselines.}
Our main competitors are the methods that can also reconstruct all three scene properties: surface geometry, materials, and illumination.
We choose NVDiffRec~\citep{NVDiffrec}, PhySG~\citep{Physg}, and IndiSG~\citep{Indirect} due to their popularity and availability of the source code. 
NVDiffRec uses tetrahedral marching to extract triangle meshes and obtains good material decomposition using a triangle-based renderer.
PhySG optimizes geometry and material information at the same time using a Spherical Gaussian representation for direct lighting and material.
IndiSG first optimizes geometry and then uses a Spherical Gaussian representation for indirect lighting to improve the quality of material reconstruction.

Apart from that, we also compared against more general surface reconstruction methods to provide additional context for our results.
For this, we use NeuS~\citep{NeuS}, Geo-NeuS~\citep{geo-neus}, and NeRO~\citep{nero}.
NeuS is a popular implicit surface reconstruction method that achieves strong results without relying on extra data.
Geo-NeuS improves upon NeuS by using additional point cloud supervision, obtained from structure from motion (SfM)~\citep{colmap_sfm}.
NeRO is capable of reconstructing reflective objects effectively.
In addition, we also compare with Ref-NeRF~\citep{Refnerf}, which considers material decomposition, but due to modeling geometry using density function, it has difficulty extracting smooth geometry.


\noindent\textbf{Evaluation metrics.}
We use the official evaluation protocol to compute Chamfer distance for the DTU dataset and also use Chamfer distance for the SK3D dataset.
We utilize the PSNR metric, to quantitatively evaluate the quality of rendering, material, and illumination. 
We follow IndiSG~\citep{Indirect} and employ masks to compute the PSNR metric in the foreground to evaluate the quality of materials and rendering.
See Appx~\apref{ap:implementation_details} for implementation details.

\begin{table*}[t]
\vspace{-0.2cm}   
\caption{Quantitative results in terms of Chamfer distance on the scenes with high specularities on DTU~\citep{DTU} and SK3D~\citep{SK3D}. Red and orange indicate the best and second-best algorithms. Geo-NeuS is separated because of the additional 3D point cloud (PC) for supervision. Our method achieves state-of-the-art surface reconstruction results on both glossy (this table) and regular (see Table~\ref{tab:compare_nero_dtu_tab}) scenes.}
\vspace{-0.2cm}  
\centering
\tiny
\resizebox{\linewidth}{!}{
\setlength{\tabcolsep}{1.4mm}{\input{tables/chamfer_distance}}
}
\label{tab:chamfer_distance}
\end{table*}

\begin{table}[h]
\caption{Quantitative results in terms of Chamfer distance on DTU.}
\vspace{-0.2cm}  
\centering
\tiny
\resizebox{0.8\linewidth}{!}
{\setlength{\tabcolsep}{0.5mm}
    \begin{tabular}{ccccccccccccccccc}
    \toprule
    DTU & 24 & 37 & 40 & 55 & 63 & 65 & 69 & 83 \\
    \midrule
    NeuS & 1.00 & 1.37 & 0.93 & 0.43 & 1.01 & 0.65 & \textbf{0.57} & 1.48 \\
    NeRO & 1.10 & 1.13 & 1.26 & 0.46 & 1.32 & 1.93 & 0.71 & 1.61 \\
    Ours & \textbf{0.82} & \textbf{1.05} & \textbf{0.85} & \textbf{0.40} & \textbf{0.99} & \textbf{0.59} & 0.60 &\textbf{1.44} \\  
    \toprule
    DTU & 97 & 105 & 106 & 110 & 114 & 118 & 122 & \textbf{Mean} \\
    \midrule
    NeuS  & 1.21 & 0.83 & \textbf{0.52} & 1.14 & \textbf{0.35} & 0.49 & 0.54 & 0.83 \\
    NeRO & 1.47 & 1.10 & 0.70 & 1.14 & 0.39  & 0.52  & 0.57 & 1.03 \\
    Ours & \textbf{1.15} & \textbf{0.81} & 0.58  & \textbf{0.89} & 0.36 & \textbf{0.44} & \textbf{0.46} & \textbf{0.76} \\
    \bottomrule
    \end{tabular}
}
\vspace{-0.5cm}  
\label{tab:compare_nero_dtu_tab}
\end{table}

\begin{figure*}[t]
\vspace{-0.2cm}   
\centering
\setlength{\abovecaptionskip}{0.7cm}
    \begin{overpic}[width=\linewidth]{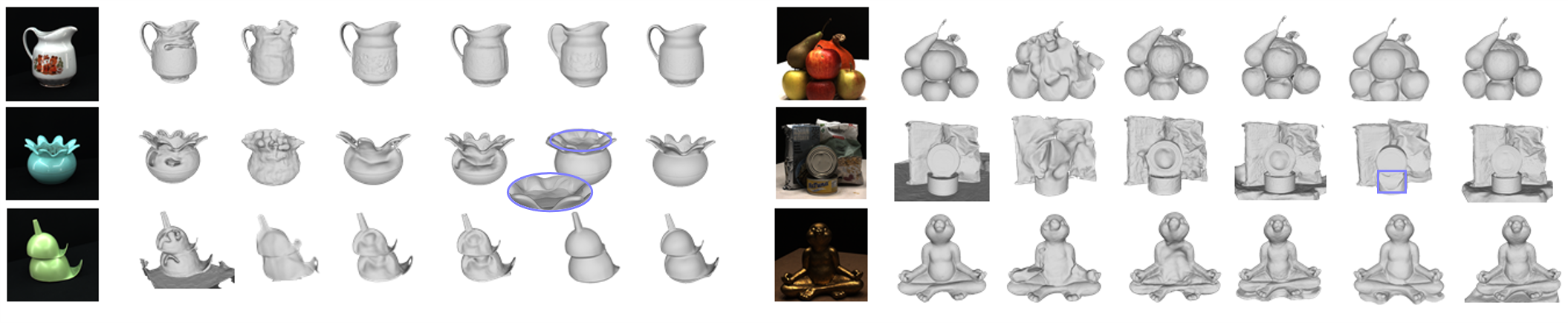}
    \put(1.6,-0.7){\fontsize{8pt}{0}\selectfont\color{black}{Image}}
    \put(8.5,-0.7){\fontsize{8pt}{0}\selectfont\color{black}{Geo-NeuS}}
    \put(16.4,-0.7){\fontsize{8pt}{0}\selectfont\color{black}{PhySG}}
    \put(22.3,-0.7){\fontsize{8pt}{0}\selectfont\color{black}{IndiSG}}
    \put(29.5,-0.7){\fontsize{8pt}{0}\selectfont\color{black}{NeuS}}
    \put(36,-0.7){\fontsize{8pt}{0}\selectfont\color{black}{NeRO}}
    \put(42.7,-0.7){\fontsize{8pt}{0}\selectfont\color{black}{Ours}}
    \put(50.4,-0.7){\fontsize{8pt}{0}\selectfont\color{black}{Image}}
    \put(57,-0.7){\fontsize{8pt}{0}\selectfont\color{black}{Geo-NeuS}}
    \put(65.2,-0.7){\fontsize{8pt}{0}\selectfont\color{black}{PhySG}}
    \put(72.7,-0.7){\fontsize{8pt}{0}\selectfont\color{black}{IndiSG}}
    \put(80,-0.7){\fontsize{8pt}{0}\selectfont\color{black}{NeuS}}
    \put(87.3,-0.7){\fontsize{8pt}{0}\selectfont\color{black}{NeRO}}
    \put(95.2,-0.7){\fontsize{8pt}{0}\selectfont\color{black}{Ours}}

    \put(-1.2,2.2){\begin{turn}{90}\fontsize{7pt}{0}\selectfont\color{black}{Funnel}\end{turn}}
    \put(-1.2,9.5){\begin{turn}{90}\fontsize{7pt}{0}\selectfont\color{black}{Pot}\end{turn}}
    \put(-1.2,15.8){\begin{turn}{90}\fontsize{7pt}{0}\selectfont\color{black}{Jug}\end{turn}}
    \put(48,1.8){\begin{turn}{90}\fontsize{7pt}{0}\selectfont\color{black}{DTU110}\end{turn}}
    \put(48,8.5){\begin{turn}{90}\fontsize{7pt}{0}\selectfont\color{black}{DTU97}\end{turn}}
    \put(48,14.9){\begin{turn}{90}\fontsize{7pt}{0}\selectfont\color{black}{DTU63}\end{turn}}
    \end{overpic}
\vspace{-0.7cm}
\caption{Qualitative results for DTU (left) and SK3D (right).}
\label{fig:surface_dtu_sk3d}
\vspace{-0.3cm}
\end{figure*}
\begin{figure*}[t]
\vspace{-0.8cm}   
\centering
\setlength{\abovecaptionskip}{0.7cm}
    \begin{overpic}[width=\linewidth]{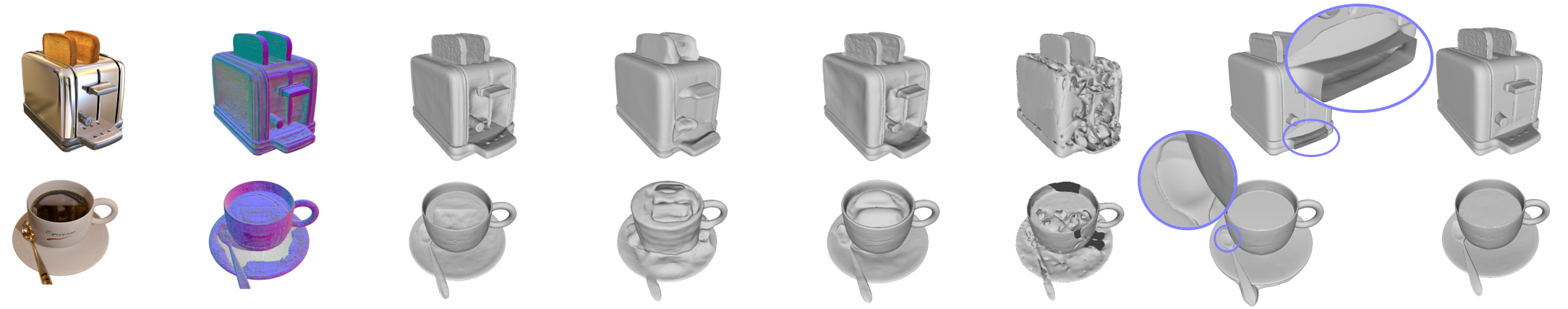}
    \put(1,-0.7){\fontsize{8pt}{0}\selectfont\color{black}{GT image}}
    \put(13.5,-0.7){\fontsize{8pt}{0}\selectfont\color{black}{Ref-NeRF}}
    \put(28,-0.7){\fontsize{8pt}{0}\selectfont\color{black}{NeuS}}
    \put(41,-0.7){\fontsize{8pt}{0}\selectfont\color{black}{PhySG}}
    \put(54,-0.7){\fontsize{8pt}{0}\selectfont\color{black}{IndiSG}}
    \put(65,-0.7){\fontsize{8pt}{0}\selectfont\color{black}{NVDiffrec}}
    \put(79,-0.7){\fontsize{8pt}{0}\selectfont\color{black}{NeRO}}
    \put(94,-0.7){\fontsize{8pt}{0}\selectfont\color{black}{Ours}}

    \put(-1,3){\begin{turn}{90}\fontsize{7pt}{0}\selectfont\color{black}{DTU110}\end{turn}}
    \put(-1,12){\begin{turn}{90}\fontsize{7pt}{0}\selectfont\color{black}{DTU97}\end{turn}}

    \end{overpic}
    \vspace{-0.8cm}
\caption{Qualitative results for the Shiny dataset(toaster and coffee).}
\label{fig:surface_shiny_part}
\vspace{0.3cm}   
\end{figure*}

\begin{figure}[t]
\vspace{-0.2cm}   
\centering{
  \begin{overpic}[width=\linewidth]{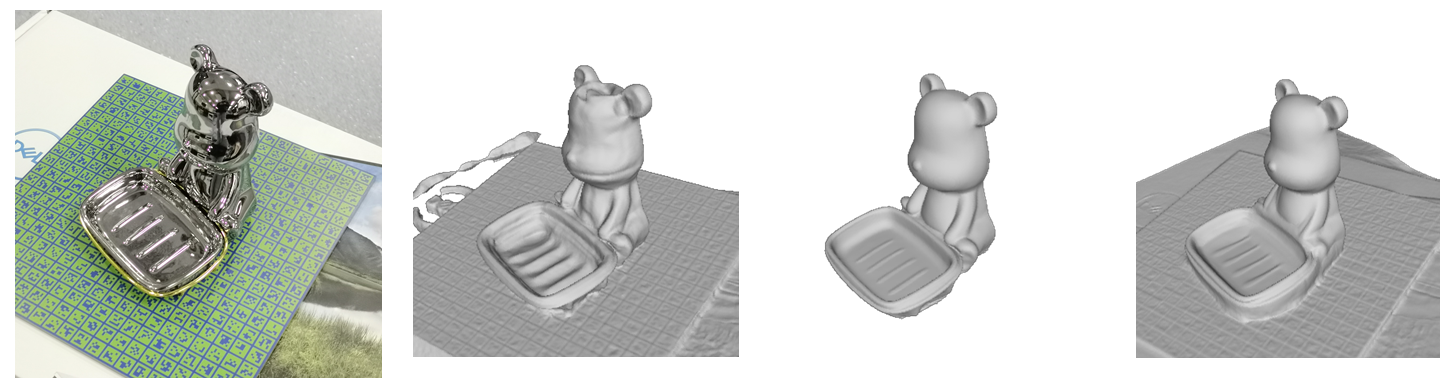}
    \put(9,-2){\fontsize{8pt}{0}\selectfont\color{black}{Image}}
    \put(37,-2){\fontsize{8pt}{0}\selectfont\color{black}{NeuS}}
    \put(61,-2){\fontsize{8pt}{0}\selectfont\color{black}{NeRO}}
    \put(86,-2){\fontsize{8pt}{0}\selectfont\color{black}{Ours}}

    \put(10.5,-6){\fontsize{8pt}{0}\selectfont\color{black}{CD}}
    \put(36,-6){\fontsize{8pt}{0}\selectfont\color{black}{0.0074}}
    \put(60.6,-6){\fontsize{8pt}{0}\selectfont\color{black}{0.0033}}
    \put(85,-6){\fontsize{8pt}{0}\selectfont\color{black}{0.0034}}
    \end{overpic}
    \vspace{-0.15cm}   
    \caption{Qualitative results for Glossy dataset.}
    \label{fig:bear}
  }
  \vspace{-0.5cm}
\end{figure}

\begin{table*}[t]
\vspace{-0.5cm}   
\caption{Quantitative PSNR results on IndiSG~\citep{Indirect} dataset for IndiSG and our method. ``SAI'' refers to specular albedo improvement. Red and orange indicate the best and second-best algorithms.}
\vspace{-0.2cm}  
\centering
\tiny
\resizebox{\linewidth}{!}{
    \setlength{\tabcolsep}{0.5mm}
    {\input{tables/nerfsyn_var}}
}
\label{tab:PSNR_IndiSG_dataset}
\vspace{-0.3cm}   
\end{table*}

\begin{figure*}[t]
\vspace{-0.5cm}   
\centering
    \begin{overpic}[width=\linewidth]{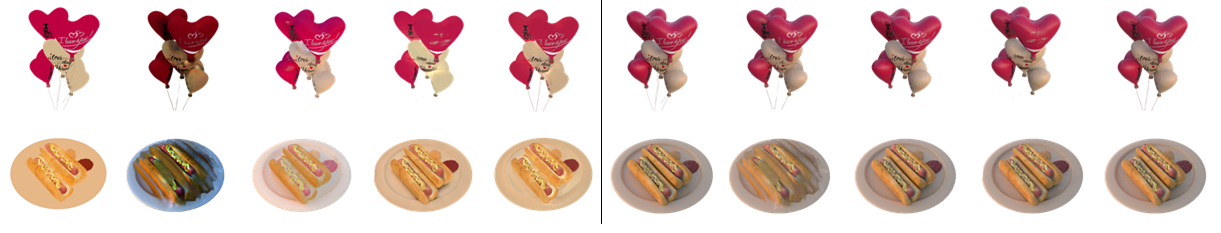}
    \put(1.4,-0.5){\fontsize{8pt}{0}\selectfont\color{black}{GT albedo}}
    \put(12,-0.5){\fontsize{8pt}{0}\selectfont\color{black}{PhySG}}
    \put(21,-0.5){\fontsize{8pt}{0}\selectfont\color{black}{NVDiffrec}}
    \put(32.4,-0.5){\fontsize{8pt}{0}\selectfont\color{black}{IndiSG}}
    \put(43.2,-0.5){\fontsize{8pt}{0}\selectfont\color{black}{Ours}}
    \put(51.5,-0.5){\fontsize{8pt}{0}\selectfont\color{black}{GT image}}
    \put(62,-0.5){\fontsize{8pt}{0}\selectfont\color{black}{PhySG}}
    \put(71,-0.5){\fontsize{8pt}{0}\selectfont\color{black}{NVDiffrec}} 
    \put(83,-0.5){\fontsize{8pt}{0}\selectfont\color{black}{IndiSG}} 
    \put(94,-0.5){\fontsize{8pt}{0}\selectfont\color{black}{Ours}}

    \put(-1.2,3){\begin{turn}{90}\fontsize{8pt}{0}\selectfont\color{black}{Hotdog}\end{turn}}
    \put(-1.2,13){\begin{turn}{90}\fontsize{8pt}{0}\selectfont\color{black}{Baloons}\end{turn}}
    \end{overpic}
    \vspace{-0.5cm}
\caption{Qualitative results on IndiSG dataset (Balloons and Hotdog) in terms of albedo reconstruction (left) and rendering quality (right).}  
\vspace{-0.2cm}  
\label{fig:decomp_indi_part}
\end{figure*}

\subsection{Surface reconstruction quality}

We first demonstrate quantitative results in terms of Chamfer distance.
We provide the numerical results for IndiSG, PhySG, NeuS, Geo-NeuS, and NeRO for comparison.
NVDiffrec struggles with glossy surface reconstruction as we subsequently verify qualitatively in \figref{fig:surface_shiny_part}. 
First, we list quantitative results on SK3D and DTU in \tabref{tab:chamfer_distance}.
As shown in the table, our approach achieves the best and the second-best performance on the SK3D and DTU scenes with glossy surfaces. Geo-NeuS achieves better performance on the DTU dataset because the additional sparse 3D point clouds generated by structure from motion (SfM) for supervising the SDF network are accurate. Our approach can also incorporate the components of Geo-NeuS based on extra data, and the surface reconstruction quality will be further improved as shown in appendix~\apref{ap:additional_experiments} (refer to it as Appx henceforth). 
However, on the SK3D scenes with more reflective surfaces, these sparse 3D points cannot be generated accurately by SfM, leading to poor surface reconstruction by Geo-NeuS.
In contrast, our approach can reconstruct glossy surfaces on both SK3D and DTU without any explicit geometry information.
IndiSG and PhySG share the same surface reconstruction method, but PhySG optimizes it together with the materials, while IndiSG freezes the underlying SDF after its initial optimization.
Compared with IndiSG, PhySG cannot optimize geometry and material information well simultaneously on real-world acquired datasets with complex lighting and materials.
We demonstrate large improvements over IndiSG and PhySG on both SK3D and DTU. NeRO focuses on highlight scenes, but our approach significantly outperforms it in both SK3D and DTU glossy scenes.
Additional Chamfer distance comparisons for the other DTU scenes are in \tabref{tab:compare_nero_dtu_tab}. 

We further demonstrate the qualitative comparison results on glossy scenes from the SK3D and DTU datasets in \figref{fig:surface_dtu_sk3d}. It can be seen that although Geo-NeuS has better quantitative evaluation metrics on DTU, it loses some of the fine details, such as the small dents on the metal can in DTU 97. By visualizing the results of the SK3D dataset, we can validate that our method can reconstruct glossy surfaces without explicit geometric supervision. 
For NeRO, even in scenarios with a simple background and straightforward geometry, it still tends to lose certain details of the dataset and fills in shadowed areas (Pot and Jug).
Furthermore, NeRO not only struggles to accurately restore detailed information but also fails to address the negative impact of partial highlights on geometry (DTU97).
More DTU results are in Appx~\nolinkfigref{fig:compare_nero_dtu_fig}.

\begin{figure}[t]
    \vspace{-0.3cm}
    \centering{
  \begin{overpic}[width=\linewidth]{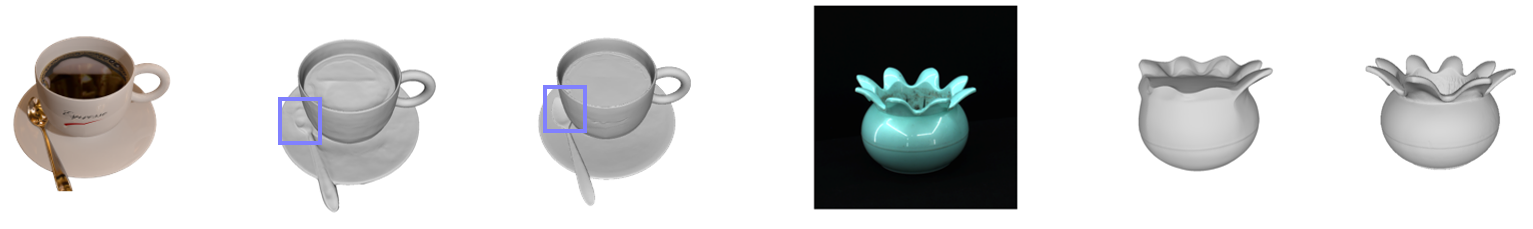}
    \put(2.2,-1){\fontsize{7pt}{0}\selectfont\color{black}{Image}}
    \put(17,-1){\fontsize{7pt}{0}\selectfont\color{black}{Ref-NeuS}}
    \put(37,-1){\fontsize{7pt}{0}\selectfont\color{black}{Ours}}
    \put(56,-1){\fontsize{7pt}{0}\selectfont\color{black}{Image}}
    \put(72.5,-1){\fontsize{7pt}{0}\selectfont\color{black}{Ref-NeuS}}
    \put(91.3,-1){\fontsize{7pt}{0}\selectfont\color{black}{Ours}}
    \put(3.5,-5){\fontsize{7pt}{0}\selectfont\color{black}{CD}}
    \put(20,-5){\fontsize{7pt}{0}\selectfont\color{black}{1.02}}
    \put(37.5,-5){\fontsize{7pt}{0}\selectfont\color{black}{0.85}}
    \put(57.5,-5){\fontsize{7pt}{0}\selectfont\color{black}{CD}}
    \put(75,-5){\fontsize{7pt}{0}\selectfont\color{black}{18.90}}
    \put(91,-5){\fontsize{7pt}{0}\selectfont\color{black}{1.54}}
    \end{overpic}
    \vspace{-0.2cm}
    \caption{Comparison with Ref-NeuS.}
    \label{fig:ref-neus}
  }
    \vspace{-0.2cm}
\end{figure}

We further show qualitative results for surface reconstruction compared with NeuS, Ref-NeRF, IndiSG, PhySG, and NVDiffrec on the Shiny dataset, which is a synthetic dataset with glossy surfaces.  From \figref{fig:surface_shiny_part}, we can observe that NeuS is easily affected by highlights, and the geometry reconstructed by Ref-NeRF has strong noise. PhySG is slightly better than IndiSG on the Shiny synthetic dataset with jointly optimizing materials and surfaces, such as toaster and car scenes, but still can not handle bright reflections.
NeRO cannot accurately reconstruct the spoon in the coffee scene and performed poorly in the toaster bottom panel. The results of other scenarios in the Shiny dataset are shown in Appx~\nolinkfigref{fig:surface_shiny}.
Our method can produce clean glossy surfaces without being affected by the issues caused by highlights. Overall, our approach demonstrates superior performance in surface reconstruction, especially on glossy surfaces. 
Additionally, we compare our method with Ref-NeuS~\cite{Refneus} in~\figref{fig:ref-neus}. Ref-NeuS’s approach is easily affected by complex specular highlights, leading to suboptimal results on models in Shiny and SK3D with complex highlights. 

For completeness, we also conducted experiments on the Glossy dataset~\citep{nero}, in which NeRO~\citep{nero} excels. In \figref{fig:bear}, we compared the reconstruction results of the bear scene with NeuS and NeRO, while other scenes are shown in Appx~\nolinkfigref{fig:compare_nero_glossy_fig}. It can be seen that our method and NeRO achieved comparable results on the Glossy dataset. Additional Chamfer distance metrics for the other scenes of the Glossy dataset are presented in Appx~\nolinktabref{tab:compare_nero_glossy_tab}.

\subsection{Material reconstruction and rendering quality}

In \tabref{tab:PSNR_IndiSG_dataset}, we evaluate the quantitative results in terms of PSNR metric for material and illumination reconstruction on the IndiSG dataset compared with PhySG, NVDiffrec, and IndiSG.
For completeness, we also compare to the case where the specular albedo improvement was not used in Stage 3 (See in Eq.~\ref{eq:spec_improvement}). Regarding diffuse albedo, although NVDiffrec showed a slight improvement over us in the balloons scene, we achieved a significant improvement over NVDiffrec in the other three scenes.
Our method achieved the best results in material reconstruction.
Moreover, our method achieves the best results in illumination quality without using the specular albedo improvement.
Additionally, our method significantly outperforms other methods in terms of rendering quality and achieves better appearance synthesis results.
We present the qualitative results of material reconstruction in \figref{fig:decomp_indi_part}, which shows that our method has better detail capture compared to IndiSG, PhySG, and NVDiffrec, such as the text on the balloon.
\nolinkfigref{fig:decomp_shiny} in Appx. also demonstrates the material decomposition effectiveness of our method on Shiny datasets with glossy surfaces.

In addition to the synthetic datasets with ground truth decomposed materials, we also provide qualitative results on real-world captured datasets (SK3D and DTU) in Appx~\nolinkfigref{fig:decomp_dtu_sk3d} and show the effectiveness of material factorization. 
To offer a more detailed presentation of the reconstruction quality across continuous viewpoints, we include videos of diffuse albedo, indirect illumination, light visibility, and rendering for three different scenes in the supplementary materials.
Furthermore, we perform relighting for these three scenes and provide videos to assess the relighting quality.

\begin{figure}[t]
\vspace{-0.2cm}   
\centering{
  \begin{overpic}[width=\linewidth]{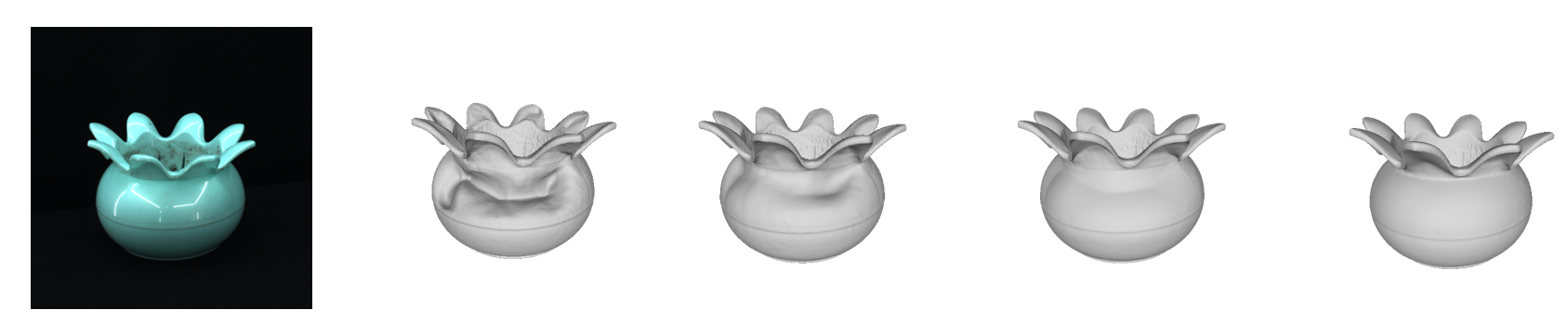}
    \put(6.5,-3){\fontsize{8pt}{0}\selectfont\color{black}{Image}}
    \put(8,-8){\fontsize{8pt}{0}\selectfont\color{black}{CD}}
    \put(28.8,-3){\fontsize{8pt}{0}\selectfont\color{black}{NeuS}}
    \put(29.5,-8){\fontsize{8pt}{0}\selectfont\color{black}{2.09}}
    \put(43.5,-3){\fontsize{8pt}{0}\selectfont\color{black}{$\Lvol$ + $\Lsur^{\uppercase\expandafter{\romannumeral1}}$}}
    \put(48.2,-8){\fontsize{8pt}{0}\selectfont\color{black}{2.01}}
    \put(64.2,-3){\fontsize{8pt}{0}\selectfont\color{black}{$\Lvol$ + $\Lvol^{\uppercase\expandafter{\romannumeral2}}$}}
    \put(68.8,-8){\fontsize{8pt}{0}\selectfont\color{black}{1.91}}
    \put(89.2,-3){\fontsize{8pt}{0}\selectfont\color{black}{Ours}}
    \put(89.5,-8){\fontsize{8pt}{0}\selectfont\color{black}{1.88}}
    \end{overpic}
    \vspace{0.05cm}   
    \caption{Ablation study for different reconstruction strategies. }
  }
  \vspace{-0.3cm}
\end{figure}

\vspace{0.1cm}   
\subsection{Ablation study}\label{sec:experiments:ablation}
\noindent\textbf{Surface reconstruction.}
To validate our surface reconstruction strategy in Stage 1, we selected the Pot scene from SK3D and ablated the method the following way.
``$\Lvol$ + $\Lsur^{\uppercase\expandafter{\romannumeral1}}$'' means that we only use volume rendering and surface rendering MLPs for surface reconstruction, without decomposing material information into diffuse and specular components.
``$\Lvol$ + $\Lvol^{\uppercase\expandafter{\romannumeral2}}$'' means we use two volume reconstructions where one of them is split into diffuse and specular components.
Just using ``$\Lvol^{\uppercase\expandafter{\romannumeral2}}$'' like Ref-NeRF to split diffuse and specular components fails to reconstruct the correct surface due to inaccurate normal vectors in the reflection direction computation.
We provide the quantitative (Chamfer distance) and qualitative results of different frameworks in \figref{fig:sur_vol_pot}.
It can be seen that synchronizing the volume color and the color on the surface point has a certain effect in suppressing concavities, but still cannot meet the requirements for complex glossy surfaces with strong reflections.
Using volume rendering to decompose diffuse and specular components can result in excessive influence from non-surface points, which still causes small concavities.
When using our loss ``$\Lvol$ ($\Lvol^{\uppercase\expandafter{\romannumeral1}}$) + $\Lsur$ ($\Lsur^{\uppercase\expandafter{\romannumeral2}}$)''
, we can achieve reconstruction results without concavities. More qualitative results are shown in Appx~\nolinkfigref{fig:ablation_more_res}.

We also implement an ablation study on the concurrent work UniSDF~\citep{wang2023unisdf} (coming code), which introduces a learnable parameter to weight the volume rendering (low-reflectance) and the specular component of volume rendering. We compared this method with ours and found that, although both approaches effectively decompose the surface, our diffuse (low-reflectance) component aligns more closely with reality with fewer specular components. This indicates that our method achieves a better decomposition.

\begin{figure}[t]
\centering{
  \begin{overpic}[width=\linewidth]{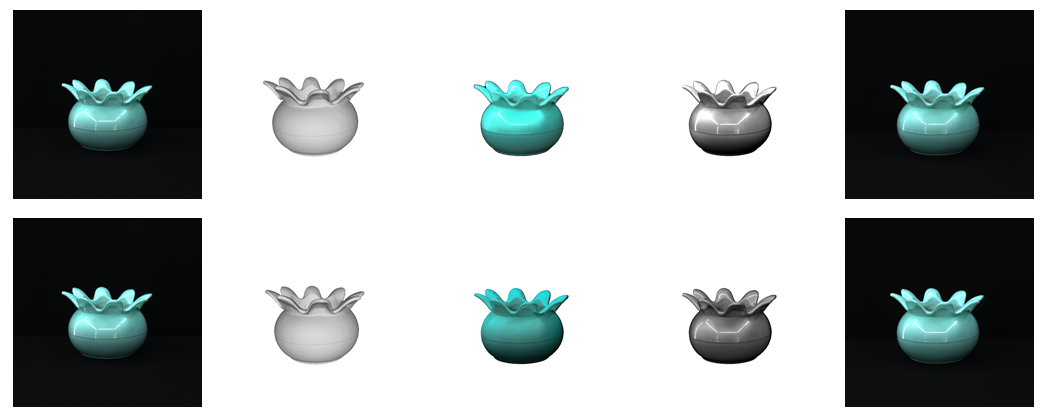}
    \put(6,-3){\fontsize{8pt}{0}\selectfont\color{black}{Image}}
    \put(26.8,-3){\fontsize{8pt}{0}\selectfont\color{black}{mesh}}
    \put(45.3,-3){\fontsize{8pt}{0}\selectfont\color{black}{diffuse}}
    \put(40.7,-6){\fontsize{8pt}{0}\selectfont\color{black}{(low specular)}}
    \put(64.8,-3){\fontsize{8pt}{0}\selectfont\color{black}{specular
}}
    \put(80,-3){\fontsize{8pt}{0}\selectfont\color{black}{rendered image
}}
     \put(-3,6.5){\begin{turn}{90}\fontsize{6pt}{0}\small\color{black}{Ours}\end{turn}}
    \put(-3,23){\begin{turn}{90}\fontsize{6pt}{0}\small\color{black}{UniSDF}\end{turn}}
    
    \end{overpic}
    \vspace{-0.2cm}   
    \caption{Comparison with UniSDF.}
    \label{fig:unisdf}
  }
  \vspace{-0.6cm}
\end{figure}

\noindent\textbf{Materials and illumination.}
We conduct an ablation study on the different components we proposed by evaluating their material and lighting performance on a complex scene, the hotdog, as shown in \tabref{tab:ablation_table}.
``SI'' refers to surface improvement, which means using networks to jointly synthesize volume color and decomposed surface color. 
``VI'' stands for visibility improvement, which involves continuous light visibility supervision based on the SDF. ``SAI'' refers to specular albedo improvement, which incorporates specular albedo into the BRDF of Spherical Gaussians.
We compare different settings in terms of diffuse albedo, roughness, appearance synthesis, and illumination. We use IndiSG as a reference and find that introducing volume rendering can improve the accuracy of material and lighting reconstruction. When the surface has no defects, further performing the surface improvement will enhance the quality of roughness and rendering but may cause a decrease in lighting reconstruction quality. Making the visibility supervision continuous improves the reconstruction of diffuse albedo, roughness, and lighting, but it also affects rendering quality. Introducing specular albedo can greatly improve roughness and rendering quality but negatively affect lighting reconstruction quality.
We further show qualitative results in Appx~\nolinkfigref{fig:ablation_about_hotdog} to highlight the importance of taking into account various factors, which enhance the reconstruction of materials and lighting.
\begin{table}[!t]
\vspace{-6pt}
    \centering
    \caption{Ablation study for materials and illumination decomposition in terms of PSNR. ``Alb'' is ``diffuse albedo'', ``Rough'' is ``roughness'', ``Rend'' is ``appearance'', and ``Illu'' is ``illumination''. }
    \vspace{-5pt} 
    \resizebox{\linewidth}{!}{
    \begin{threeparttable}[b]
        \input{tables/ablation_hotdog}
    \end{threeparttable}
    }
    \label{tab:ablation_table}
    \vspace{-0.5cm}  
\end{table}
In stage 3, if we do not consider indirect illumination during the training process, the predicted results for rendering, material, and lighting will all experience a decline. The qualitative results are shown in Appx~\nolinkfigref{fig:ablation_indiLgt_fig}. The specific PSNR metrics can be found in \tabref{tab:PSNR_IndiSG_dataset}

\section{Conclusions}
In this work, we propose Factored-NeuS, a novel approach to inverse rendering that factors geometry, material, and lighting from multiple views.
Our first contribution is to simultaneously synthesize the appearance, diffuse radiance, and specular radiance during surface reconstruction. The combination of volume and decoupled surface rendering allows the geometry to be unaffected by glossy highlights.
Our second contribution is to train networks to estimate reflectance albedo and learn a visibility function supervised by continuous values based on the SDF, so that our method is capable of better decomposing material and lighting.
Experimental results show that our method surpasses the state-of-the-art in both geometry reconstruction quality and material reconstruction quality.
A future research direction is how to effectively decompose materials for fine structures.

\section*{Acknowledgements}
We would like to acknowledge support from the National Natural Science Foundation of China (62202076) and the KAUST - Center of Excellence for Generative AI, under award number 5940, and the NTGC-AI program.


{
    \small
    \bibliographystyle{cvpr2025/ieeenat_fullname}

\input{main.bbl}
}

\clearpage

\appendix
\clearpage

\section{Preliminaries}
\label{ap:preliminaries}

\textbf{Volume rendering}. 
The radiance $C$ of the pixel corresponding to a given ray $\ray(t) = \origin + t\raydir$ at the origin $\bm o \in \R^3$ towards direction $\bm d \in \Ss^2$ is calculated using the volume rendering equation, which involves an integral along the ray with boundaries $t_n$ and $t_f$ ($t_n$ and $t_f$ are parameters to define the near and far clipping plane).
This calculation requires the knowledge of the volume density $\sigma$ and directional color $\bm{c}$ for each point within the volume.
\begin{equation}
C(\ray) = \int_{{t_n}}^{{t_f}} {T(t)\sigma (\ray(t)){\bf{c}}(\ray(t),\raydir)} dt
\label{eq:rendering_equ}
\end{equation}
The volume density $\sigma$ is used to calculate the accumulated transmittance $T(t)$:
\begin{equation}
T(t)=\exp \left(-\int_{t_n}^{t_f} \sigma(\ray_s) d s \right)
\end{equation}
It is then used to compute a weighting function $\weight(t) = T(t)\sigma (\ray(t))$ to weigh the sampled colors along the ray $\ray(t)$ to integrate into radiance $C(\ray)$.

\noindent\textbf{Surface rendering}. 
The radiance $L_o(\coord, \diro)$ reflected from a surface point $\coord$ in direction $\diro = -\raydir$ is an integral of bidirectional reflectance distribution function (BRDF) and illumination over half sphere $\Omega$, centered at normal $\normal$ of the surface point $\coord$:
\begin{equation}
L_o(\coord, \diro)=\int_\Omega L_{i}(\coord, \diri) f_r(\coord, \diri, \diro)(\diri \cdot \normal) \mathrm{d} \diri
\label{eq:surface_rendering}
\end{equation}
where $L_{i}(\coord, \diri)$ is the illumination on $\coord$ from the incoming light direction $\diri$, and $f_r$ is BRDF, which is the proportion of light reflected from direction $\diri$ towards direction $\diro$ at the point $\coord$.

\section{Implementation details}
\label{ap:implementation_details}

Our full model is composed of several MLP networks, each one of them having a width of 256 hidden units unless otherwise stated.
In Stage 1, the SDF network $\sdf$ is composed of 8 layers and includes a skip connection at the 4-th layer, similar to NeuS~\citep{NeuS}.
The input 3D coordinate $\coord$ is encoded using positional encoding with 6 frequency scales.
The diffuse color network $\netdiff$ utilizes a 4-layer MLP, while the input surface normal $\normal$ is positional-encoded using 4 scales.
For the specular color network $\netspec$, a 4-layer MLP is employed, and the reflection direction $\dirr$ is also positional-encoded using 4 frequency scales.
In the first stage, we exclusively focus on decomposing the highlight (largely white) areas. To reduce the complexity of considering color, we assume that the specular radiance is in grayscale and only consider changes in brightness. We can incorporate color information in later stages to obtain a more detailed specular reflection model. Like NeuS, the background is modeled by NeRF++.

In Stage 2, the light visibility network $\netvis$ has 4 layers.
To better encode the input 3D coordinate $\coord$, positional encoding with 10 frequency scales is utilized. The input view direction $\omega_i$ is also positional-encoded using 4 scales.
The indirect light network $\netind$ in stage 2 comprises 4 layers.

In stage 3, the encoder part of the BRDF network consists of 4 layers, and the input 3D coordinate is positional-encoded using 10 scales.
The output latent vector $\latent$ has 32 dimensions, and we impose a sparsity constraint on the latent code $\latent$, following IndiSG~\citep{Indirect}.
The decoder part of the BRDF network is a 2-layer MLP with a width of 128, and the output has 4 dimensions, including the diffuse albedo $\diffalb \in \R^3$ and roughness $r \in \R$.
Finally, the specular albedo network $\netspecalb$ uses a 4-layer MLP, where the input 3D coordinate $\bm{x}$ is positional-encoded using 10 scales, and the input reflection direction $\dirr$ is positional-encoded using 4 scales. 

The learning rate for all three stages begins with a linear warm-up from 0 to $5 \times 10^{-4}$ during the first 5K iterations.
It is controlled by the cosine decay schedule until it reaches the minimum learning rate of $2.5 \times 10^{-5}$, which is similar to NeuS.
The weights $\Lsurcoef$ for the surface color loss are set for 0.1, 0.6, 0.6, 0.6 and 0.01 for DTU, SK3D, Shiny, Glossy, and the IndiSG dataset, respectively. For all datasets, the Fresnel value f in the rendering equation is set to 0.02.
We train our model for 300K iterations in the first stage, which takes 11 hours in total.
For the second and third stages, we train for 40K iterations, taking around 1 hour each.
The training was performed on a single NVIDIA RTX 4090 GPU.

\section{Training strategies of stage 1}
\label{ap:training_strategies}

In our training process, we define three loss functions, namely volume radiance loss $\Lvol$, surface radiance loss $\Lsur$, and regularization loss $\Lreg$. 
The volume radiance loss $\Lvol$ is measured by calculating the $\Loss_1$ distance between the ground truth colors $\colorgt$ and the volume radiances $\colorvol$ of a subset of rays $\rayset$, which is defined as follows.
\begin{equation}
    \Lvol = \frac{1}{|\rayset|}\sum_{\ray \in \rayset} \|\colorvol_{\ray} - \colorgt_{\ray}\|_1
\end{equation}
The surface radiance loss $\Lsur$ is measured by calculating the $\Loss_1$ distance between the ground truth colors $\colorgt$ and the surface radiances $\colorsur$.
During the training process, only a few rays have intersection points with the surface. We only care about the set of selected rays $\rayset'$, which satisfies the condition that each ray exists point whose SDF value is less than zero and not the first sampled point.
The loss is defined as follows.
\begin{equation}
    \Lsur=\frac{1}{|\rayset'|}\sum_{\ray \in \rayset'} \|\colorsur_{\ray} - \colorgt_{\ray}\|_1
\end{equation}
$\Lreg$ is an Eikonal loss term on the sampled points.
Eikonal loss is a regularization loss applied to a set of sampling points $X$, which is used to constrain the noise in signed distance function (SDF) generation.
\begin{equation}
    \Lreg = \frac{1}{|\coordset|}\sum_{\coord \in \coordset}(\| \nabla \sdf(\coord)\|_2 -1)^2
\end{equation}
We use weights $\Lsurcoef$ and $\Lregcoef$ to balance the impact of these three losses.
The overall training weights are as follows.
\begin{equation}
    \Loss = \Lvol + \Lsurcoef\Lsur + \Lregcoef\Lreg
\end{equation}

\section{Details of stage 2}
\label{ap:details_s2}

At this stage, we focus on predicting the lighting visibility and indirect illumination of a surface point $\coord$ under different incoming light direction $\diri$ using the SDF in the first stage. Therefore, we need first to calculate the position of the surface point $\coord$. In stage one, we have calculated two sampling points $\ray(t_{\izero-1}), \ray(t_\izero)$ near the surface. As Geo-NeuS~\citep{geo-neus}, we weigh these two sampling points to obtain a surface point $\coord$ as follows.
\begin{equation}
    \coord=\dfrac{\sdf(\ray(t_{\izero-1}))\ray(t_\izero)-\sdf(\ray(t_\izero))\ray(t_{\izero-1})}{\sdf(\ray(t_{\izero-1}))-\sdf(\ray(t_\izero))}
\end{equation}
\noindent\textbf{Learning lighting visibility.}
Visibility is an important factor in shadow computation.
It calculates the visibility of the current surface point $\coord$ in the direction of the incoming light $\diri$.
Path tracing of the SDF is commonly used to obtain a binary visibility (0 or 1) as used in IndiSG~\citep{Indirect}, but this kind of visibility is not friendly to network learning.
Inspired by NeRFactor~\citep{Nerfactor}, we propose to use an integral representation with the continuous weight function $\weight(t)$ (from 0 to 1) for the SDF to express light visibility.
Specifically, we establish a neural network $\netvis: (\coord, {\diri}) \mapsto \vis$, that maps the surface point $\coord$ and incoming light direction $\diri$ to visibility, and the ground truth value of light visibility is obtained by integrating the weights $\weight_i$ of the SDF of sampling points along the incoming light direction and can be expressed as follows. 
\begin{equation}
\vis^{gt} = 1 - \sum_{i=1}^n\weight_i
\end{equation}
The weights of the light visibility network are optimized by minimizing the loss between the calculated ground truth values and the predicted values of a set of sampled incoming light directions $\Omega_i \subset \Ss^2$.
This pre-integrated technique can reduce the computational burden caused by the integration for subsequent training.
\begin{equation} 
\Loss_{\text{vis}} = \frac{1}{|\Omega_i|}\sum_{\bm{\omega} \in \Omega_i} {\|\vis_{\bm{\omega}} - \vis_{\bm{\omega}}^{\text{gt}}\|_1}
\end{equation}

\noindent\textbf{Learning indirect illumination.}
Indirect illumination refers to the light that is reflected or emitted from surfaces in a scene and then illuminates other surfaces, rather than directly coming from a light source, which contributes to the realism of rendered images.
Following IndiSG~\citep{Indirect}, we parameterize indirect illumination $I(\coord,{\diri})$ via $K_i=24$ Spherical Gaussians (SGs) as follows.
\begin{equation}
I(\coord,\diri)=\sum_{k=1}^{K_i} I_k (\diri \ |\  \sgaxisind_k(\coord), \sgsharpind_k(\coord), \sgamplind_k(\coord))
\end{equation}
where $\sgaxisind_k(\coord) \in \Ss^2$, $\sgsharpind_k(\coord) \in \R_+$, and $\sgamplind_k(\coord) \in \R^3$ are the lobe axis, sharpness, and amplitude of the $k$-th Spherical Gaussian, respectively. 
For this, we train a network $\netind: \coord \mapsto \{\sgaxisind_k, \sgsharpind_k, \sgamplind_k\}_{k=1}^{K_i}$ that maps the surface point $\coord$ to the parameters of indirect light SGs. 
Similar to learning visibility, we randomly sample several directions $\diri$ from the surface point $\coord$ to obtain (pseudo) ground truth $I^{\text{gt}}(\coord,\diri)$.
Some of these rays have intersections $\coord^{\prime}$ with other surfaces, thus, $\diri$ is the direction pointing from $\coord$ to $\coord^{\prime}$.
We query our proposed color network $\netcolor$ to get the (pseudo) ground truth indirect radiance $I^{\text{gt}}(\coord,\diri)$ as follows.  
\begin{equation}
I^{\text{gt}}(\coord,\diri)=\netcolor(\coord^{\prime},  {\normal}^{\prime},\diri, \feat)
\end{equation}
where ${\normal}^{\prime}$ is the normal on the point $\coord^{\prime}$. We also use $\Loss_1$ loss to train the network.
\begin{equation}
    \mathcal{L}_{\text{ind}}=\frac{1}{|M|}\sum_{m\in M} {\|I(\coord,\boldsymbol{\omega}_{m})-I_m^{\text{gt}}(\coord,\boldsymbol{\omega}_{m})\|_1}
\end{equation}

\section{Details of stage 3}
\label{ap:details_s3}
The combination of light visibility and illumination SG is achieved by applying a ratio to the lobe amplitude of the output SG, while preserving the center position of the SG. We randomly sample $K_s = 32$ directions within the SG lobe and compute a weighted average of the visibility with different directions.

\begin{equation}
\begin{aligned}
     & \vis(\coord, \diri) \otimes E_k (\diri \ |\ \sgaxisenv_k, \sgsharpenv_k, \sgamplenv_k) \\
     \approx & E_k (\diri \ |\ \sgaxisenv_k, \sgsharpenv_k, \frac{\sum_{s=1}^{K_s} E_k(\boldsymbol{\omega}_s)\vis(\coord,\boldsymbol{\omega}_s)}{\sum_{s=1}^{K_s} E_k(\boldsymbol{\omega}_s)}\sgamplenv_k) 
\end{aligned}
\end{equation}

Here, we offer intuitive explanations for why the incorporation of specular albedo in the model results in a decrease in lighting prediction.
The increase in the model's complexity is the primary reason. Specular albedo introduces a more detailed modeling of surface reflection characteristics, requiring additional parameters and learning capacity. This raises the difficulty of training the model, potentially resulting in overfitting or training instability, thereby affecting the accurate prediction of lighting.

\section{Additional results}
\label{ap:more_results}
\subsection{Additional results for the main text}
\begin{figure*}[t]
\centering
    \begin{overpic}[width=\linewidth]{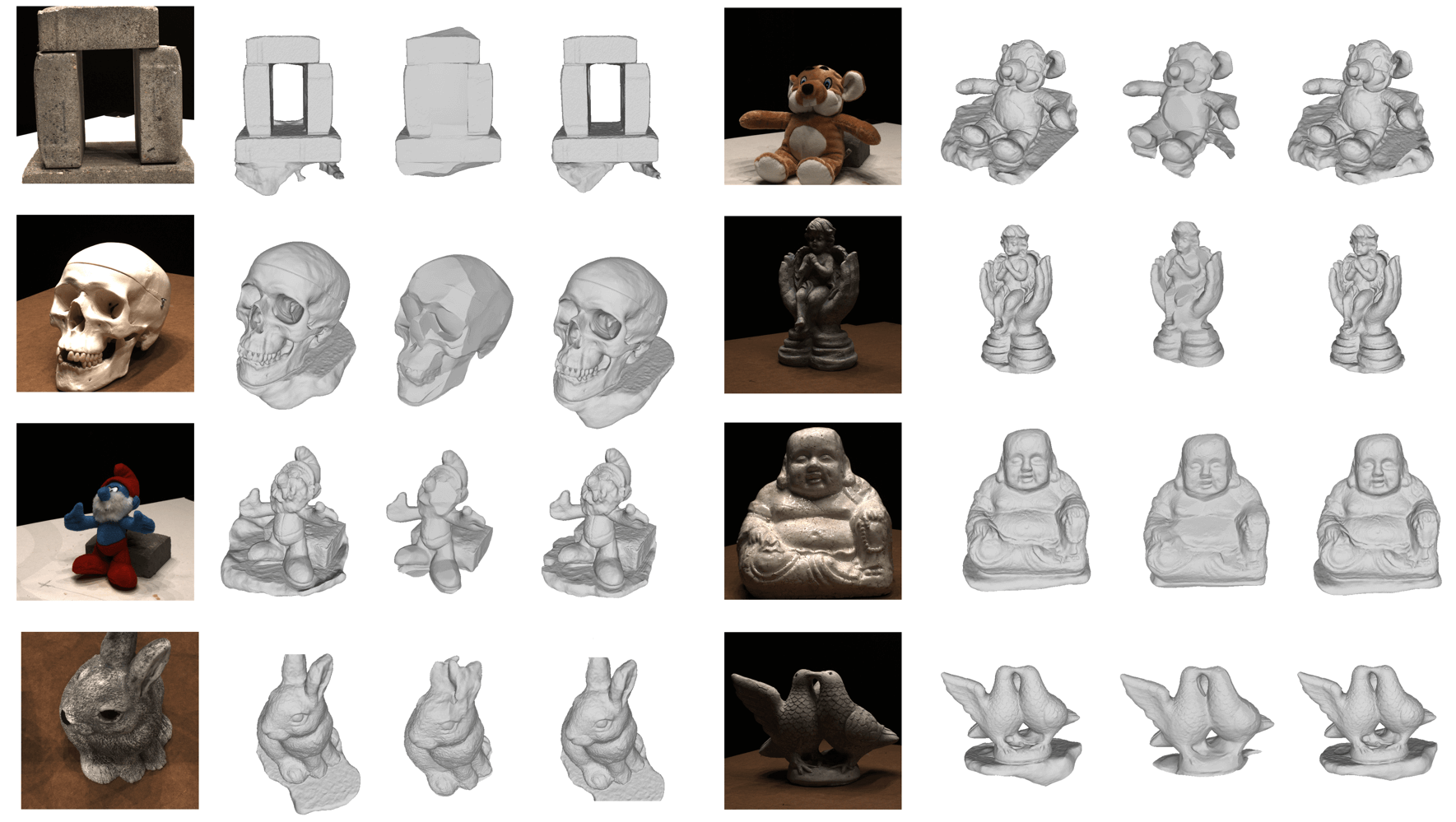}
    
    \put(4,-2){\fontsize{8pt}{0}\selectfont\color{black}{reference}}
    \put(5,-4){\fontsize{8pt}{0}\selectfont\color{black}{image}}
    
    \put(18.5,-2){\fontsize{8pt}{0}\selectfont\color{black}{NeuS}}
    
    \put(28.5,-2){\fontsize{8pt}{0}\selectfont\color{black}{NeRO}}

    \put(40.3,-2){\fontsize{8pt}{0}\selectfont\color{black}{Ours}}
    
    \put(52,-2){\fontsize{8pt}{0}\selectfont\color{black}{reference}}
    \put(53,-4){\fontsize{8pt}{0}\selectfont\color{black}{image}}
    
    \put(67.8,-2){\fontsize{8pt}{0}\selectfont\color{black}{NeuS}}
    
    \put(79.5,-2){\fontsize{8pt}{0}\selectfont\color{black}{NeRO}}
    
    \put(91.5,-2){\fontsize{8pt}{0}\selectfont\color{black}{Ours}}

    \end{overpic}
    \vspace{0.3cm}
\caption{Qualitative results on regular scenes from DTU. with NeRO and NeuS on DTU dataset. Results show that NeRO fails to address the negative impact of partial highlights on the geometry. Moreover, the presence of shadows causes NeRO to mistakenly reconstruct shadowed areas as real objects and fill them in (bricks and skull models).}
\label{fig:compare_nero_dtu_fig}
\end{figure*}

\begin{figure*}[h]
\centering
    \begin{overpic}[width=\linewidth, clip]{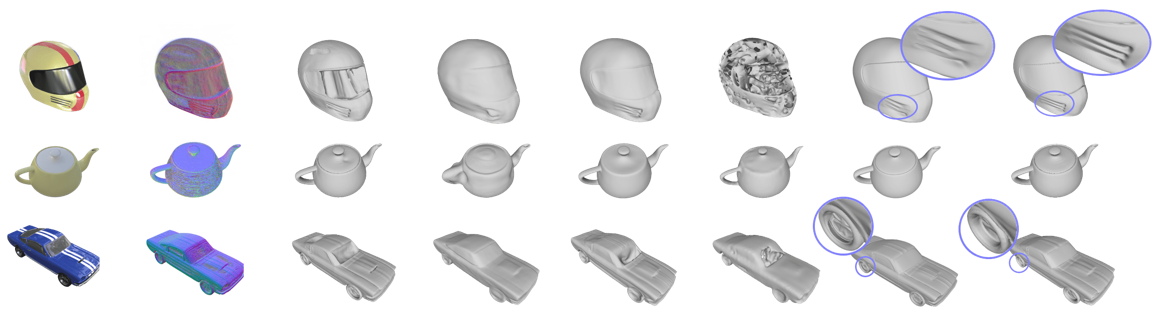}
    \put(2,-0.5){\fontsize{15pt}{0}\small\color{black}{GT image}}
    \put(14,-0.5){\fontsize{15pt}{0}\small\color{black}{Ref-NeRF}}
    \put(28,-0.5){\fontsize{15pt}{0}\small\color{black}{NeuS}}
    \put(40,-0.5){\fontsize{15pt}{0}\small\color{black}{PhySG}}
    \put(52,-0.5){\fontsize{15pt}{0}\small\color{black}{IndiSG}}
    \put(63,-0.5){\fontsize{15pt}{0}\small\color{black}{NVDiffrec}}
    \put(76,-0.5){\fontsize{15pt}{0}\small\color{black}{NeRO}}
    \put(90,-0.5){\fontsize{15pt}{0}\small\color{black}{Ours}}

    \put(-1,4.5){\begin{turn}{90}\fontsize{15pt}{0}\small\color{black}{Car}\end{turn}}
    \put(-1,11.5){\begin{turn}{90}\fontsize{15pt}{0}\small\color{black}{Teapot}\end{turn}}
    \put(-1,19.5){\begin{turn}{90}\fontsize{15pt}{0}\small\color{black}{Helmet}\end{turn}}
    \end{overpic}
\vspace{-0.4cm}
\caption{Qualitative results on other scans (helmet, teapot, and car) from the Shiny dataset~\citep{Refnerf}. NeRO exhibits defects in the dents and highlights of the helmet, as well as in the wheels of the car.}
\label{fig:surface_shiny}
\end{figure*}

\begin{figure*}[t]
\centering
    \begin{overpic}[width=\linewidth]{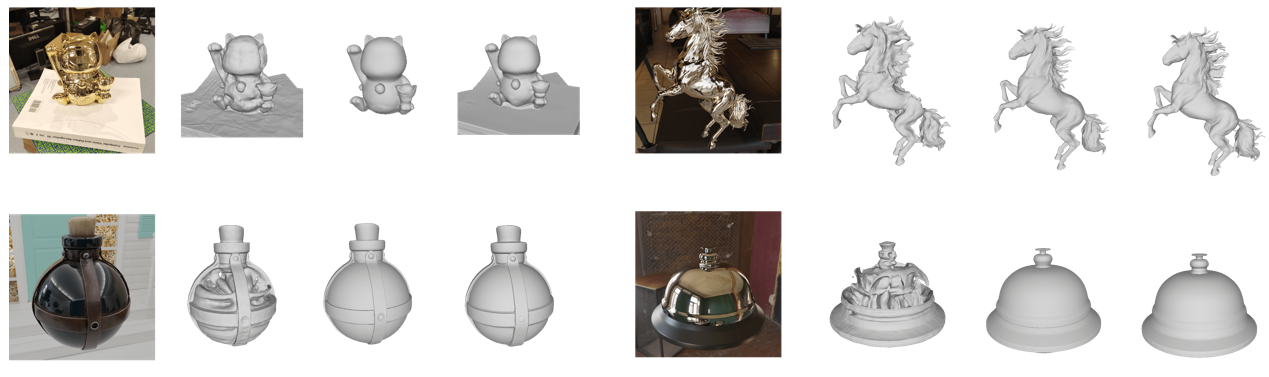}
    
    \put(2.5,-2){\fontsize{8pt}{0}\selectfont\color{black}{reference}}
    \put(3.5,-4){\fontsize{8pt}{0}\selectfont\color{black}{image}}
    
    \put(16,-2){\fontsize{8pt}{0}\selectfont\color{black}{NeuS}}
    
    \put(26.5,-2){\fontsize{8pt}{0}\selectfont\color{black}{NeRO}}

    \put(38,-2){\fontsize{8pt}{0}\selectfont\color{black}{Ours}}
    
    \put(52.2,-2){\fontsize{8pt}{0}\selectfont\color{black}{reference}}
    \put(53.2,-4){\fontsize{8pt}{0}\selectfont\color{black}{image}}
    
    \put(68,-2){\fontsize{8pt}{0}\selectfont\color{black}{NeuS}}
    
    \put(80,-2){\fontsize{8pt}{0}\selectfont\color{black}{NeRO}}
    
    \put(93,-2){\fontsize{8pt}{0}\selectfont\color{black}{Ours}}

    \end{overpic}
    \vspace{0.3cm}
\caption{Qualitative results on other scenes (bell, potion, horse, and tbell) compared with NeRO and NeuS on the Glossy dataset. }
\label{fig:compare_nero_glossy_fig}
\end{figure*}

\begin{figure*}[t]
    \centering
    \begin{overpic}[width=\linewidth]{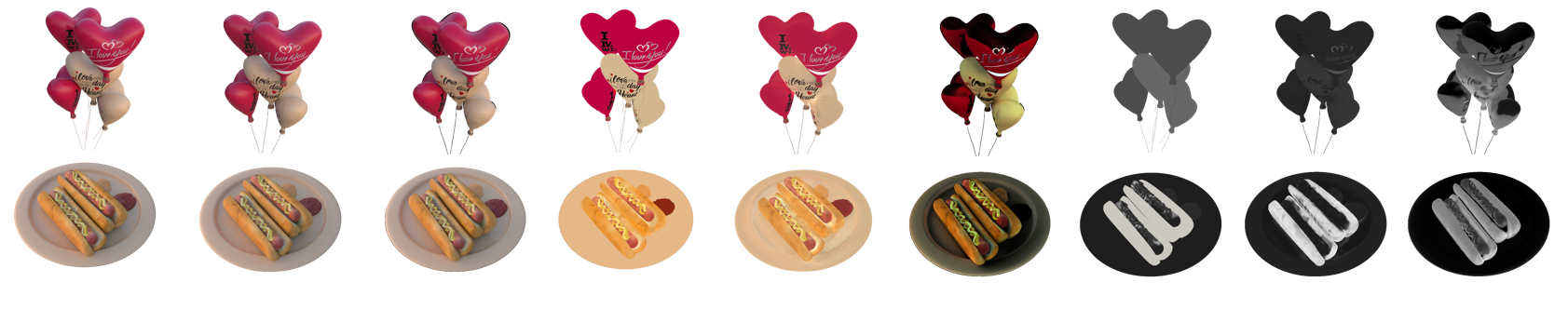}
    \put(2.5,0.2){\fontsize{8pt}{0}\selectfont\color{black}{GT Image}}
    \put(16,0.2){\fontsize{8pt}{0}\selectfont\color{black}{Ours}}
    \put(27.5,0.2){\fontsize{8pt}{0}\selectfont\color{black}{NeRO}}
    \put(36.5,0.2){\fontsize{8pt}{0}\selectfont\color{black}{GT Albedo}}
    \put(50,0.2){\fontsize{8pt}{0}\selectfont\color{black}{Ours}}
    \put(60.5,0.2){\fontsize{8pt}{0}\selectfont\color{black}{NeRO}}
    \put(70.5,0.2){\fontsize{8pt}{0}\selectfont\color{black}{GT Rough}}
    \put(82.7,0.2){\fontsize{8pt}{0}\selectfont\color{black}{Ours}}
    \put(92.7,0.2){\fontsize{8pt}{0}\selectfont\color{black}{NeRO}}
    \end{overpic}
\vspace{-0.3cm}
\caption{Qualitative comparison with NeRO in terms of material reconstruction and rendering quality.}
\label{fig:nero_compare_material}
\end{figure*}

We conduct a more in-depth comparison of our method with the already published work NeRO. 
For DTU datasets, our findings demonstrate that NeRO performs less effectively than our approach on real datasets DTU, especially in the regular scenes of DTU shown in \figref{fig:compare_nero_dtu_fig}. NeRO fails to address the negative impact of partial highlights on the geometry. For example, the highlighted region of the skull model is reconstructed as overly flat, while the Buddha model loses numerous details that should have been retained. Similar issues are also observed in the two plush toy scenes. Moreover, the presence of shadows causes NeRO to mistakenly reconstruct shadowed areas as real objects and fill them in (bricks and skull models).
\figref{fig:surface_shiny} shows three other scenes (helmet, teapot, and car) from the Shiny dataset. We have demonstrated that our method performs better than other methods on the Shiny dataset. NeRO exhibits defects in the dents and highlights of the helmet, as well as in the wheels of the car.
Furthermore, we extend our comparison to include more scenes in the glossy dataset in \figref{fig:compare_nero_glossy_fig} and \tabref{tab:compare_nero_glossy_tab}, where although NeRO performs better, our method is also capable of mitigating the impact of highlights on geometry. Our method demonstrates comparable results to NeRO. Moreover, compared to NeuS, the results show a significant improvement. Note that the real datasets include bear, bunny, coral, and vase. The rest of the others are synthetic. 
As for material representation, we adopted the spherical Gaussians to represent the Ward BRDF model~\citep{wardbrdf}, while NeRO uses the spherical harmonics to represent the Disney BRDF model~\citep{Disney}. \tabref{tab:nero_compare_material} and \figref{fig:nero_compare_material} show that we outperform NeRO in materials and rendering.

\begin{table*}[h]
\caption{Comparison with NeRO and NeuS on Glossy dataset.}
\centering
\tiny
\resizebox{0.9\linewidth}{!}
{\setlength{\tabcolsep}{0.5mm}
\begin{tabular}{ccccccccccccccc}
\toprule
Glossy & bear & bunny & coral & maneki & vase & angel & bell & cat & horse & luyu & potion & tbell & teapot & mean \\
\midrule
NeuS & 0.0074 & 0.0022  &0.0016 & 0.0091 &0.0101 &0.0035 & 0.0146 &0.0278 & 0.0053 &0.0066 &0.0393 & 0.0348 & 0.0546 & 0.0167 \\
NeRO &\textbf{0.0033} &\textbf{0.0012} & \textbf{0.0014} &\textbf{0.0024} &\textbf{0.0011} & \textbf{0.0034} &\textbf{0.0032} &  \textbf{0.0044} &\textbf{0.0049} & \textbf{0.0054} & \textbf{0.0053} & \textbf{0.0035} & \textbf{0.0037} &\textbf{0.0033} \\
Ours & 0.0034  & 0.0017& \textbf{0.0014} & 0.0027 & 0.0023 & \textbf{0.0034} & 0.0054&  0.0059 & 0.0052 & 0.0060 & 0.0058 & \textbf{0.0035} & 0.0105 & 0.0044 \\
\bottomrule
\end{tabular}
}
\label{tab:compare_nero_glossy_tab}
\end{table*}

\begin{table}[h]
    \caption{Comparison of material rendering on IndiSG dataset.}
    \centering
    \resizebox{\linewidth}{!}{
        \begin{tabular}{lcccccc}
            \toprule
            & \multicolumn{3}{c}{Baloons} & \multicolumn{3}{c}{Hotdog} \\
            & albedo & rough & render & albedo & rough & render \\
            \midrule
            NeRO  & 14.65 & 18.91  & 23.84 & 11.54 & 18.42  & 26.95 \\
            Ours & \textbf{25.79} & \textbf{19.75}  & \textbf{33.89} & \textbf{30.72} & \textbf{23.10} & \textbf{36.71} \\
            \bottomrule
        \end{tabular}
    }
    \label{tab:nero_compare_material}
\end{table}

\begin{figure*}[t]
\centering
    \begin{overpic}[width=\linewidth]{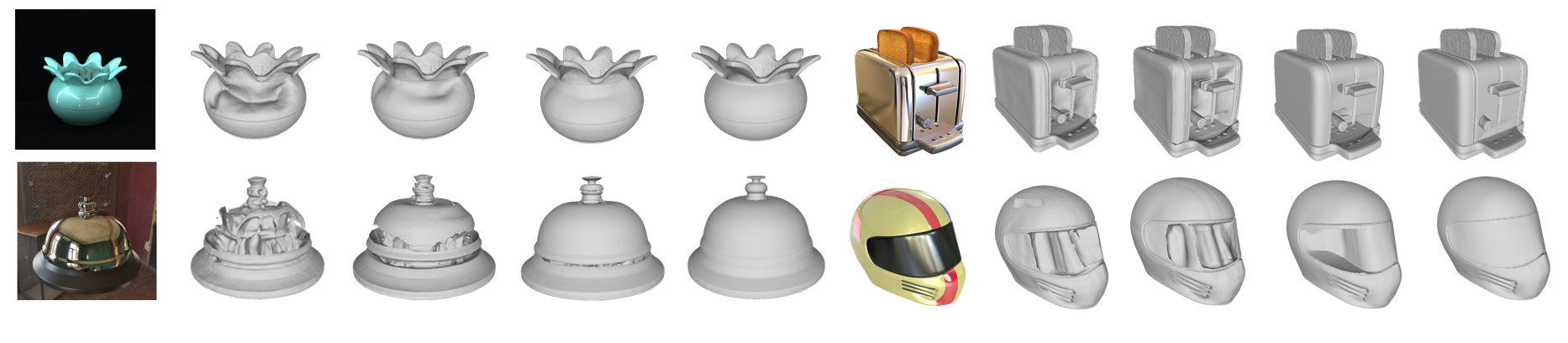}%
    
    \put(2.7,0.7){\fontsize{8pt}{0}\selectfont\color{black}{GT Image}}
    \put(15.2,0.7){\fontsize{8pt}{0}\selectfont\color{black}{NeuS}}
    \put(23.8,0.7){\fontsize{8pt}{0}\selectfont\color{black}{$\Loss_\text{vol}$ + $\Lsur^{\uppercase\expandafter{\romannumeral1}}$}}
    \put(35.1,0.7){\fontsize{8pt}{0}\selectfont\color{black}{$\Lvol$ + $\Lvol^{\uppercase\expandafter{\romannumeral2}}$}}
    \put(47.2,0.7){\fontsize{8pt}{0}\selectfont\color{black}{Ours}}
    \put(56,0.7){\fontsize{8pt}{0}\selectfont\color{black}{GT Image}}
    \put(66.6,0.7){\fontsize{8pt}{0}\selectfont\color{black}{NeuS}}
    \put(73,0.7){\fontsize{8pt}{0}\selectfont\color{black}{$\Loss_\text{vol}$ + $\Lsur^{\uppercase\expandafter{\romannumeral1}}$}}
    \put(83,0.7){\fontsize{8pt}{0}\selectfont\color{black}{$\Lvol$ + $\Lvol^{\uppercase\expandafter{\romannumeral2}}$}}
    \put(94.8,0.7){\fontsize{8pt}{0}\selectfont\color{black}{Ours}}
    
    \end{overpic}
    \vspace{-0.3cm}
\caption{Qualitative ablation evaluation for more objects.}
\label{fig:ablation_more_res}
\end{figure*}

In order to prove the effectiveness of the combination of volume rendering and decoupled surface rendering, in \figref{fig:ablation_more_res}, we additionally present the qualitative evaluation for more objects. It is evident that surface rendering is essential for decomposing diffuse and specular components, ensuring smooth reconstruction of glossy surfaces with complex reflections.

\begin{figure*}[t]
\centering
    \begin{overpic}[width=\linewidth]{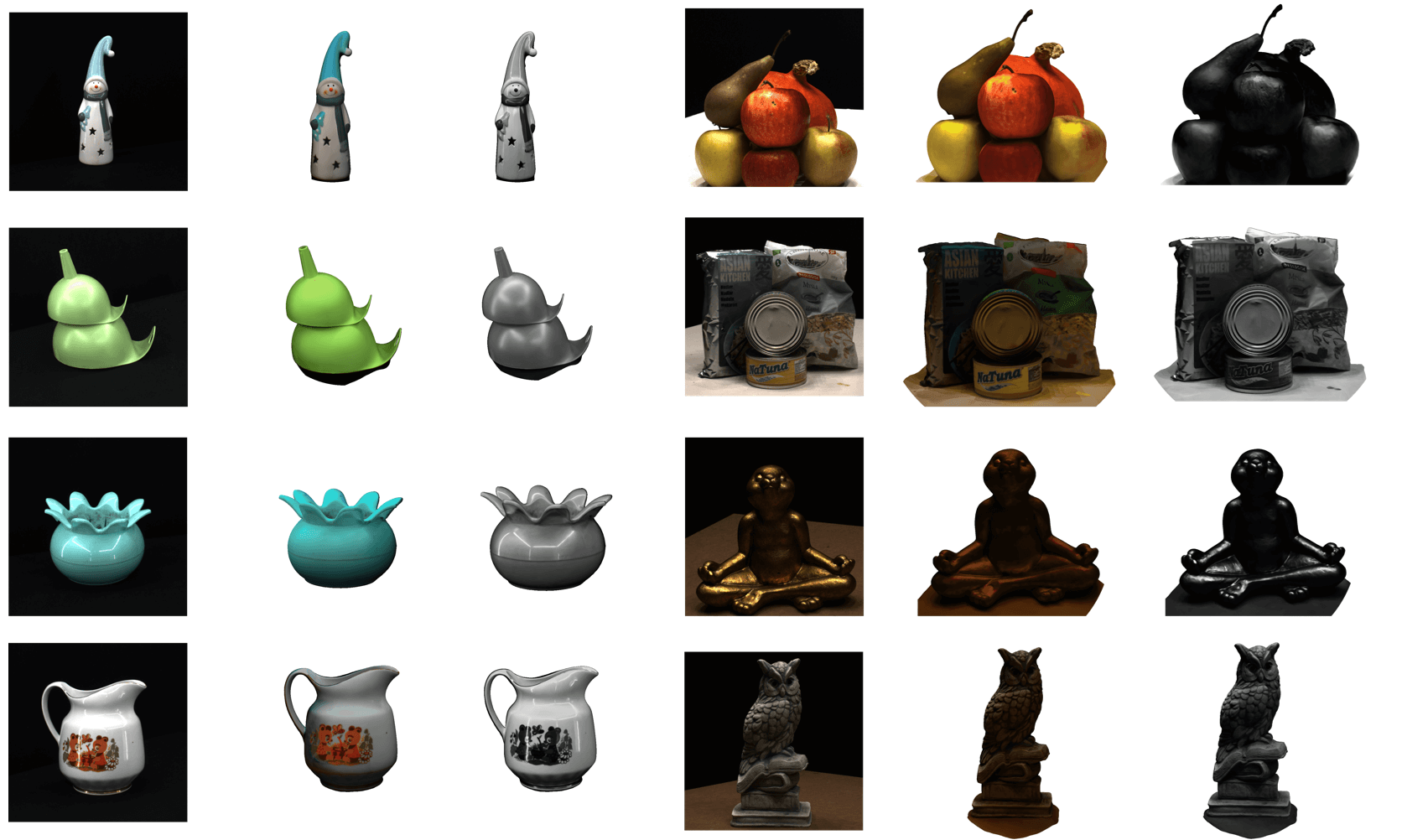}
    
    \put(3.2,-2){\fontsize{15pt}{0}\small\color{black}{GT image}}
    
    \put(22.2,-2){\fontsize{15pt}{0}\small\color{black}{diffuse}}
    \put(22.6,-4){\fontsize{15pt}{0}\small\color{black}{color}}
    
    \put(36.3,-2){\fontsize{15pt}{0}\small\color{black}{specular}}
    \put(37.4,-4){\fontsize{15pt}{0}\small\color{black}{color}}
    
    \put(52.2,-2){\fontsize{15pt}{0}\small\color{black}{GT image}}
    
    \put(70.7,-2){\fontsize{15pt}{0}\small\color{black}{diffuse}}
    \put(71.5,-4){\fontsize{15pt}{0}\small\color{black}{color}}
    
    \put(87.7,-2){\fontsize{15pt}{0}\small\color{black}{specular}}
    \put(88.7,-4){\fontsize{15pt}{0}\small\color{black}{color}}

    \end{overpic}
    \vspace{0.3cm}
\caption{Diffuse and specular decomposition results in the first stage. }
\label{fig:first_stage}
\end{figure*}

\figref{fig:decomp_indi} illustrates the qualitative results of material reconstruction on the other scenes of the IndiSG dataset, highlighting the effectiveness of our method.
For completeness, we visualize the decomposition of diffuse and specular in the first stage in \figref{fig:first_stage}. In the first stage, the decomposition of diffuse and specular is not a true BRDF model. This is because the MLP in the first stage is used solely for predicting the components of diffuse and specular reflection, rather than predicting material properties such as albedo and roughness. The decision to directly predict colors instead of material properties in the first stage serves two purposes: reducing model complexity by focusing on the direct prediction of specular reflection color, and optimizing geometry for better reconstruction. By decomposing highlights through the network in the first stage, surfaces with specular reflections can be reconstructed more effectively, demonstrated by the presence of flower pot ablation, and without encountering the concavity issues observed in other methods.

\begin{figure*}[h]
\centering
    \begin{overpic}[width=\linewidth]{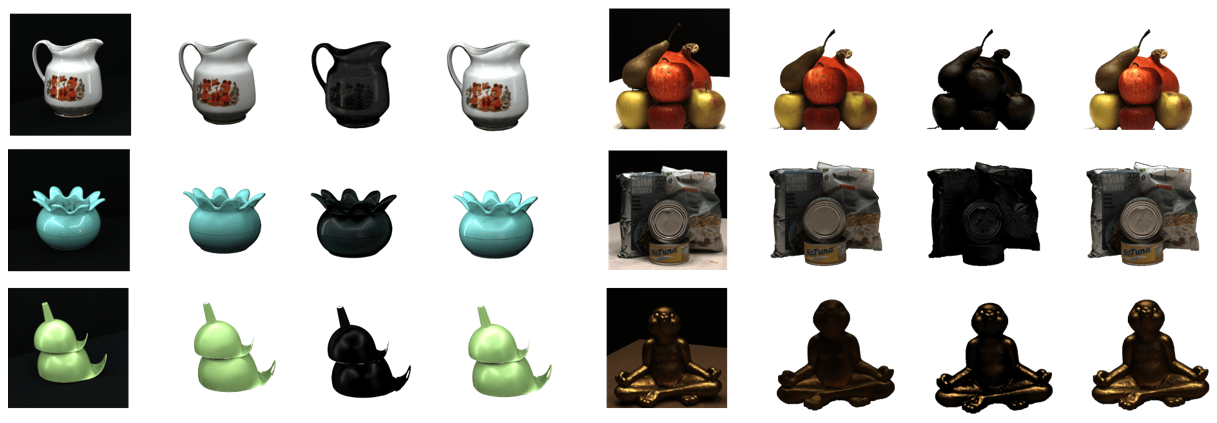}
    \put(3,-1){\fontsize{8pt}{0}\selectfont\color{black}{Image}}
    \put(17.2,-1){\fontsize{8pt}{0}\selectfont\color{black}{Ours}}
    \put(16.5,-3){\fontsize{8pt}{0}\selectfont\color{black}{Diffuse}}
    \put(28,-1){\fontsize{8pt}{0}\selectfont\color{black}{Ours}}
    \put(27,-3){\fontsize{8pt}{0}\selectfont\color{black}{Specular}}
    \put(40,-1){\fontsize{8pt}{0}\selectfont\color{black}{Ours}}
    \put(38,-3){\fontsize{8pt}{0}\selectfont\color{black}{Appearance}}
    \put(53,-1){\fontsize{8pt}{0}\selectfont\color{black}{Image}}
    \put(67.8,-1){\fontsize{8pt}{0}\selectfont\color{black}{Ours}}
    \put(67,-3){\fontsize{8pt}{0}\selectfont\color{black}{Diffuse}}
    \put(80.8,-1){\fontsize{8pt}{0}\selectfont\color{black}{Ours}}
    \put(79.5,-3){\fontsize{8pt}{0}\selectfont\color{black}{Specular}}
    \put(93.5,-1){\fontsize{8pt}{0}\selectfont\color{black}{Ours}}
    \put(91.4,-3){\fontsize{8pt}{0}\selectfont\color{black}{Appearance}}

    \put(-2,4){\begin{turn}{90}\fontsize{7pt}{0}\selectfont\color{black}{Funnel}\end{turn}}
    \put(-2,16.5){\begin{turn}{90}\fontsize{7pt}{0}\selectfont\color{black}{Pot}\end{turn}}
    \put(-2,27.8){\begin{turn}{90}\fontsize{7pt}{0}\selectfont\color{black}{Jug}\end{turn}}
    \put(48,3.5){\begin{turn}{90}\fontsize{7pt}{0}\selectfont\color{black}{DTU110}\end{turn}}
    \put(48,15){\begin{turn}{90}\fontsize{7pt}{0}\selectfont\color{black}{DTU97}\end{turn}}
    \put(48,26.5){\begin{turn}{90}\fontsize{7pt}{0}\selectfont\color{black}{DTU63}\end{turn}}
    \end{overpic}
    \vspace{0.2cm}
\caption{Qualitative results for the SK3D (left) and DTU (right) datasets in the third stage. We can observe that our method can separate the specular reflection component from the diffuse reflection component, as seen in the highlights on the apple, can, and golden rabbit. Even when faced with a higher intensity of specular reflection in SK3D, our method can separate them very well.}  
\label{fig:decomp_dtu_sk3d}
\end{figure*}

In \figref{fig:decomp_dtu_sk3d}, from the DTU data, we can observe that our method can separate the specular reflection component from the diffuse reflection component, as seen in the highlights on the apple, can, and golden rabbit. Even when faced with a higher intensity of specular reflection, as demonstrated in the example showcased in SK3D, our method excels at preserving the original color in the diffuse part and accurately separating highlights into the specular part.

\begin{figure*}[h]
\centering
    \begin{overpic}[width=\linewidth]{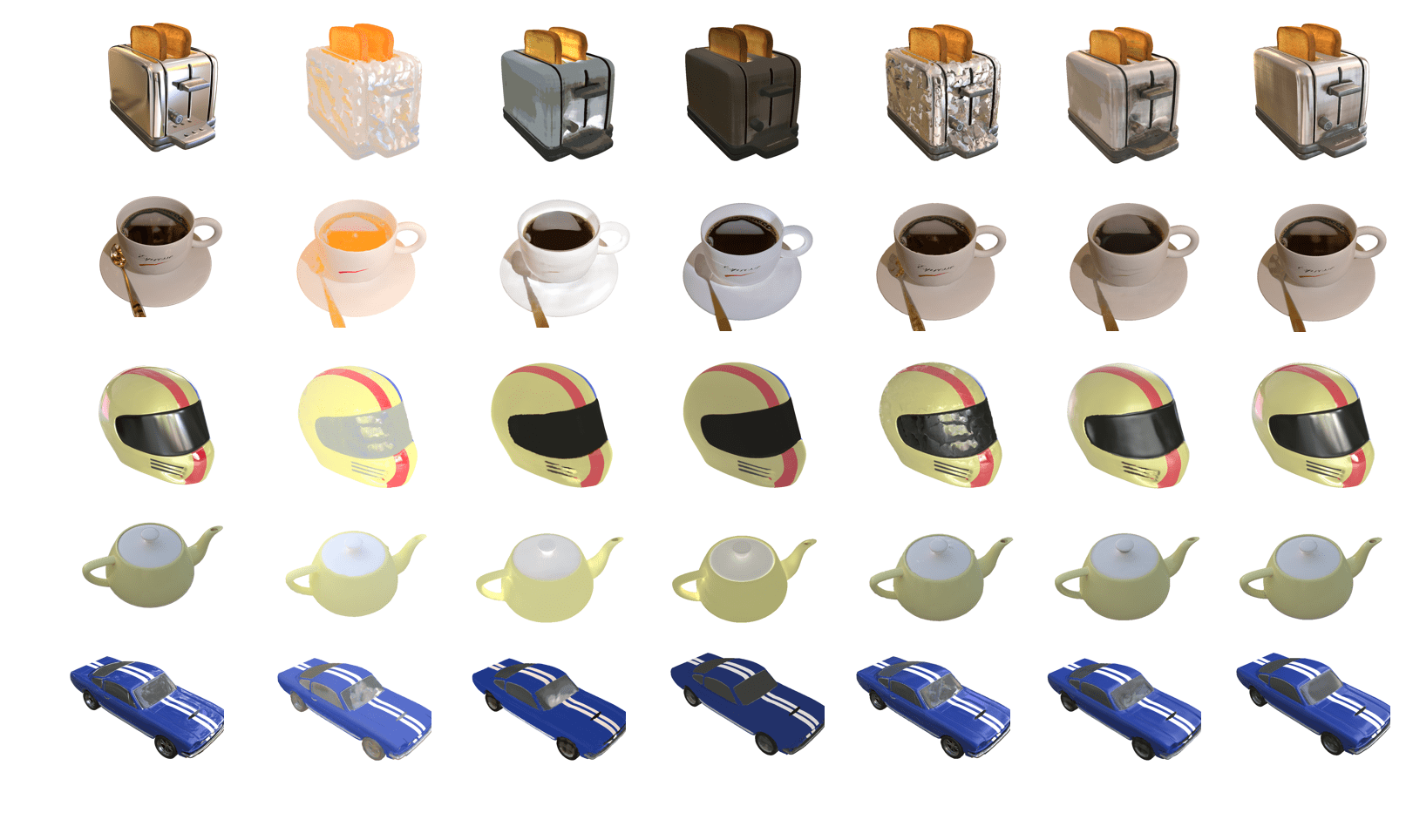}
    \put(6,1){\fontsize{15pt}{0}\small\color{black}{GT image}}
    \put(21,1){\fontsize{15pt}{0}\small\color{black}{Ref-NeRF}}
    \put(38,1){\fontsize{15pt}{0}\small\color{black}{NeuS}}
    \put(52,1){\fontsize{15pt}{0}\small\color{black}{PhySG}}
    \put(64.8,1){\fontsize{15pt}{0}\small\color{black}{IndiSG}}
    \put(77,1){\fontsize{15pt}{0}\small\color{black}{NVDiffrec}}
    \put(92,1){\fontsize{15pt}{0}\small\color{black}{Ours}}

    \put(22,-1){\fontsize{15pt}{0}\small\color{black}{Albedo}}
    \put(37.2,-1){\fontsize{15pt}{0}\small\color{black}{Albedo}}
    \put(52,-1){\fontsize{15pt}{0}\small\color{black}{Albedo}}
    \put(65,-1){\fontsize{15pt}{0}\small\color{black}{Image}}
    \put(78.2,-1){\fontsize{15pt}{0}\small\color{black}{Image}}
    \put(91.5,-1){\fontsize{15pt}{0}\small\color{black}{Image}}

    \put(0,8){\begin{turn}{90}\fontsize{15pt}{0}\small\color{black}{Car}\end{turn}}
    \put(0,17){\begin{turn}{90}\fontsize{15pt}{0}\small\color{black}{Teapot}\end{turn}}
    \put(0,28){\begin{turn}{90}\fontsize{15pt}{0}\small\color{black}{Helmet}\end{turn}}
    \put(0,39){\begin{turn}{90}\fontsize{15pt}{0}\small\color{black}{Coffee}\end{turn}}
    \put(0,50){\begin{turn}{90}\fontsize{15pt}{0}\small\color{black}{Toaster}\end{turn}}
    \end{overpic}
\vspace{-0.2cm}
\caption{Qualitative results of materials reconstruction for the Shiny dataset, where albedo refers to the diffuse albedo. The results indicate that our method can restore the original appearance with specular highlights more accurately, such as the reflections on the helmet and toaster, and has a better ability to suppress decomposition ambiguity caused by specular highlights.}
\label{fig:decomp_shiny}
\end{figure*}

In \figref{fig:decomp_shiny}, we show the diffuse albedo and rendering results of NVDiffrec, IndiSG, and our method. The rendering results indicate that our method can restore the original appearance with specular highlights more accurately, such as the reflections on the helmet and toaster compared to the IndiSG and NVDiffrec methods. The material reconstruction results show that our diffuse albedo contains less specular reflection information compared to other methods, indicating our method has a better ability to suppress decomposition ambiguity caused by specular highlights.

\begin{figure*}[t]
    \begin{overpic}[width=\linewidth]{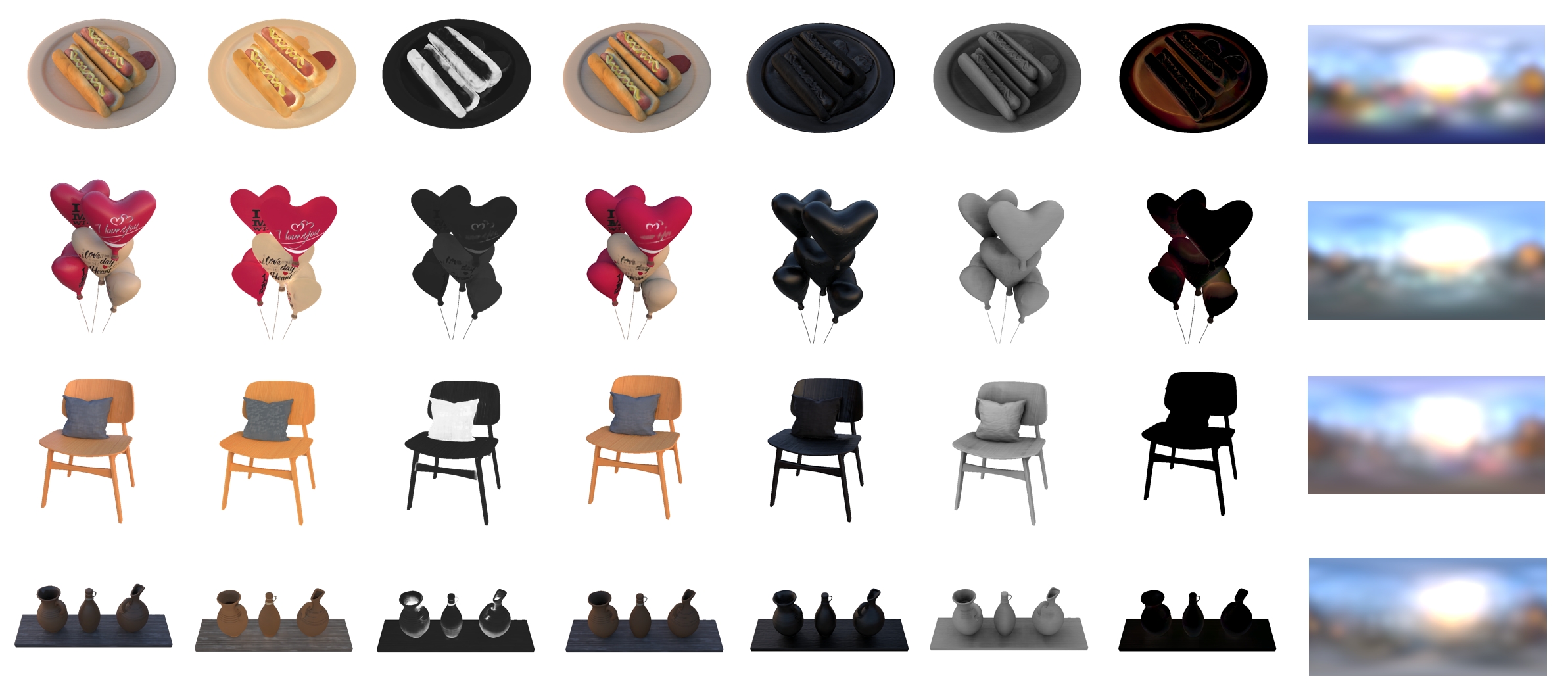}
    
    \put(2.8,0){\fontsize{8pt}{0}\selectfont\color{black}{rendering}}
    
    \put(15,0){\fontsize{8pt}{0}\selectfont\color{black}{albedo}}
    
    \put(26,0){\fontsize{8pt}{0}\selectfont\color{black}{roughness}}

    \put(39,0){\fontsize{8pt}{0}\selectfont\color{black}{diffuse}}
    \put(39.5,-2){\fontsize{8pt}{0}\selectfont\color{black}{color}}
    
    \put(50,0){\fontsize{8pt}{0}\selectfont\color{black}{specular}}
    \put(51,-2){\fontsize{8pt}{0}\selectfont\color{black}{color}}
    
    \put(63,0){\fontsize{8pt}{0}\selectfont\color{black}{light}}
    \put(62,-2){\fontsize{8pt}{0}\selectfont\color{black}{visibility}}
    
    \put(74,0){\fontsize{8pt}{0}\selectfont\color{black}{indirect}}
    \put(75,-2){\fontsize{8pt}{0}\selectfont\color{black}{light}}
    
    \put(89,-1){\fontsize{8pt}{0}\selectfont\color{black}{light}}

    \end{overpic}
    \vspace{0cm}
\caption{Visualization of all components on IndiSG dataset.}
\label{fig:all_indisg}
\end{figure*}

\begin{figure*}[t]
\centering
    \begin{overpic}[width=\linewidth]{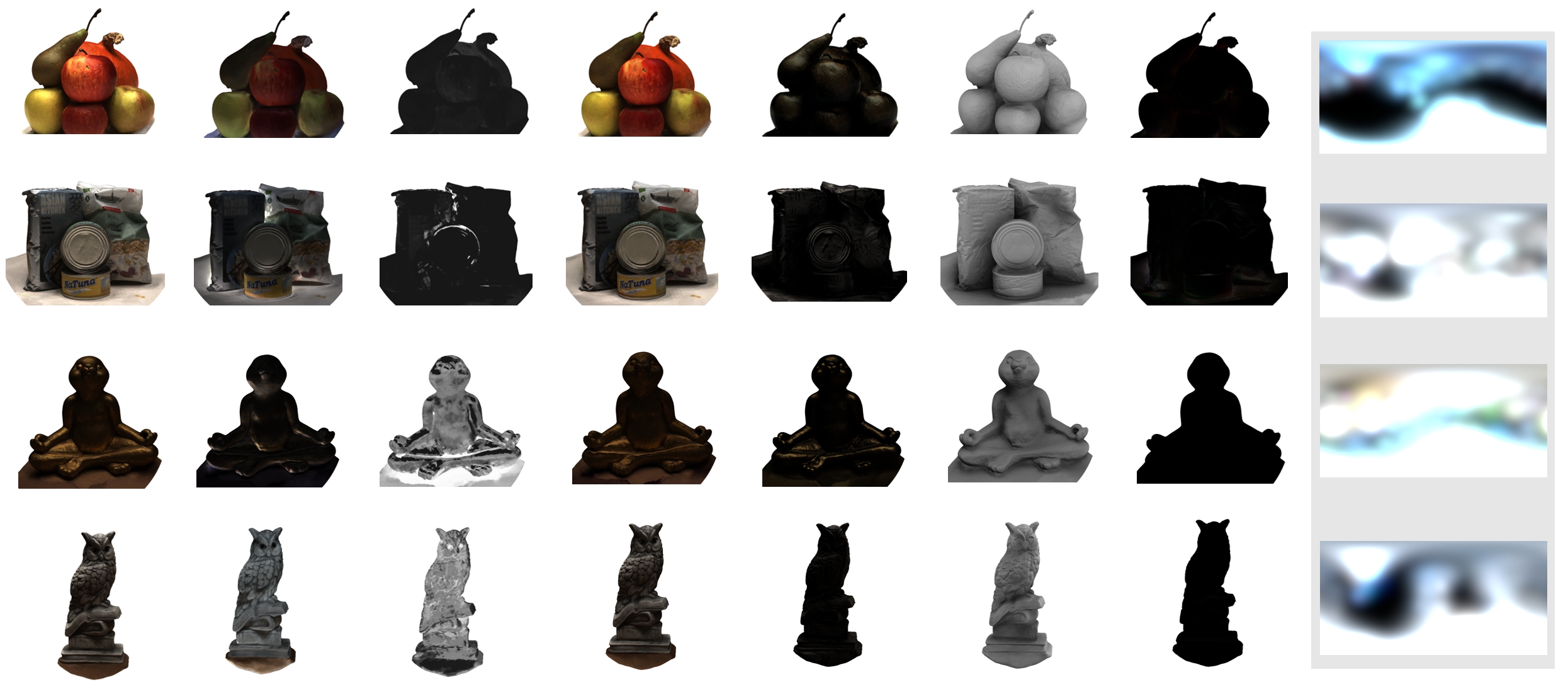}
    
    \put(2.5,-1){\fontsize{8pt}{0}\selectfont\color{black}{rendering}}
    
    \put(14.5,-1){\fontsize{8pt}{0}\selectfont\color{black}{albedo}}
    
    \put(24.5,-1){\fontsize{8pt}{0}\selectfont\color{black}{roughness}}

    \put(38,-1){\fontsize{8pt}{0}\selectfont\color{black}{diffuse}}
    \put(39,-3){\fontsize{8pt}{0}\selectfont\color{black}{color}}
    
    \put(50,-1){\fontsize{8pt}{0}\selectfont\color{black}{specular}}
    \put(51,-3){\fontsize{8pt}{0}\selectfont\color{black}{color}}
    
    \put(63,-1){\fontsize{8pt}{0}\selectfont\color{black}{light}}
    \put(61.5,-3){\fontsize{8pt}{0}\selectfont\color{black}{visibility}}
    
    \put(74,-1){\fontsize{8pt}{0}\selectfont\color{black}{indirect}}
    \put(75,-3){\fontsize{8pt}{0}\selectfont\color{black}{light}}
    
    \put(89,-1){\fontsize{8pt}{0}\selectfont\color{black}{light}}

    \end{overpic}
    \vspace{0.2cm}
\caption{Visualization of all components on DTU dataset.}
\label{fig:all_dtu}
\end{figure*}

\begin{figure*}[t]
\centering
    \begin{overpic}[width=\linewidth]{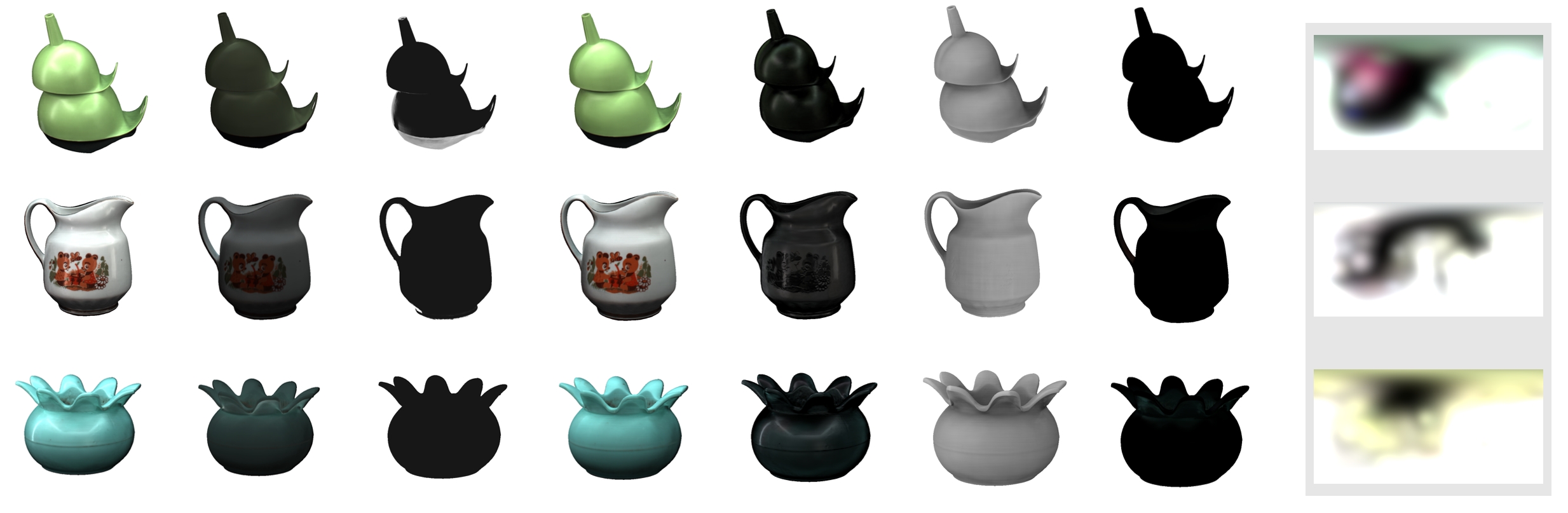}
    
    \put(1.5,-1){\fontsize{8pt}{0}\selectfont\color{black}{rendering}}
    
    \put(14.5,-1){\fontsize{8pt}{0}\selectfont\color{black}{albedo}}
    
    \put(24,-1){\fontsize{8pt}{0}\selectfont\color{black}{roughness}}

    \put(37.5,-1){\fontsize{8pt}{0}\selectfont\color{black}{diffuse}}
    \put(38.5,-3){\fontsize{8pt}{0}\selectfont\color{black}{color}}
    
    \put(49,-1){\fontsize{8pt}{0}\selectfont\color{black}{specular}}
    \put(50,-3){\fontsize{8pt}{0}\selectfont\color{black}{color}}
    
    \put(62,-1){\fontsize{8pt}{0}\selectfont\color{black}{light}}
    \put(61,-3){\fontsize{8pt}{0}\selectfont\color{black}{visibility}}
    
    \put(72.5,-1){\fontsize{8pt}{0}\selectfont\color{black}{indirect}}
    \put(74,-3){\fontsize{8pt}{0}\selectfont\color{black}{light}}
    
    \put(89,-1){\fontsize{8pt}{0}\selectfont\color{black}{light}}

    \end{overpic}
    \vspace{0.2cm}
\caption{Visualization of all components on SK3D dataset.}
\label{fig:all_sk3d}
\end{figure*}

Additionally, in \figref{fig:all_indisg}, \figref{fig:all_dtu}, and \figref{fig:all_sk3d}, we presented all components, the rendering, albedo, roughness, diffuse color, specular color, light visibility, indirect light, and environment light results for the IndiSG, DTU and SK3D datasets, respectively. An interesting observation is that our reconstructed environment maps have the capability to represent multiple direct light illuminants, as demonstrated in the DTU dataset.

\begin{figure*}[t]
\centering
    \begin{overpic}[width=\linewidth]{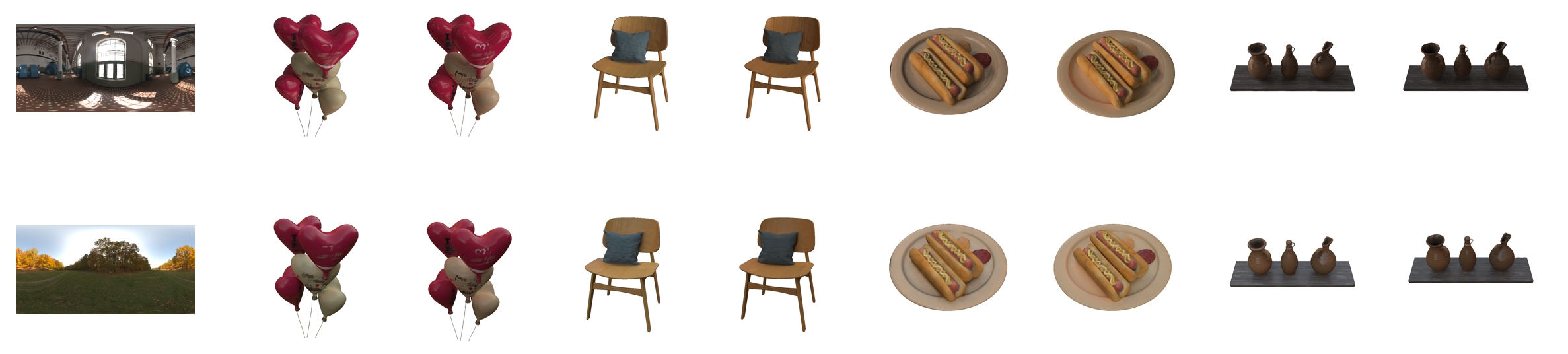}
    
    \put(5,-2){\fontsize{8pt}{0}\selectfont\color{black}{light}}
    
    \put(17.5,-2){\fontsize{8pt}{0}\selectfont\color{black}{IndiSG}}
    
    \put(27.5,-2){\fontsize{8pt}{0}\selectfont\color{black}{Ours}}

    \put(37.5,-2){\fontsize{8pt}{0}\selectfont\color{black}{IndiSG}}
    
    \put(48,-2){\fontsize{8pt}{0}\selectfont\color{black}{Ours}}
    
    \put(58.5,-2){\fontsize{8pt}{0}\selectfont\color{black}{IndiSG}}
    
    \put(69,-2){\fontsize{8pt}{0}\selectfont\color{black}{Ours}}
    
    \put(80,-2){\fontsize{8pt}{0}\selectfont\color{black}{IndiSG}}
    
    \put(92,-2){\fontsize{8pt}{0}\selectfont\color{black}{Ours}}

    \end{overpic}
    \vspace{0.2cm}
\caption{Relighting comparison with IndiSG.}
\label{fig:compare_relighting}
\end{figure*}

In \figref{fig:compare_relighting}, we additionally showcase the visualization results of relighting compared with the IndiSG method.
IndiSG and ours yield different predictions for material, resulting in variations in the relighting results, but the relighting results generated by our method exhibit richer details. 
Our method demonstrates the practical utility employed in the relighting scenarios.

\begin{figure*}[h]
\centering
    \begin{overpic}[width=\linewidth]{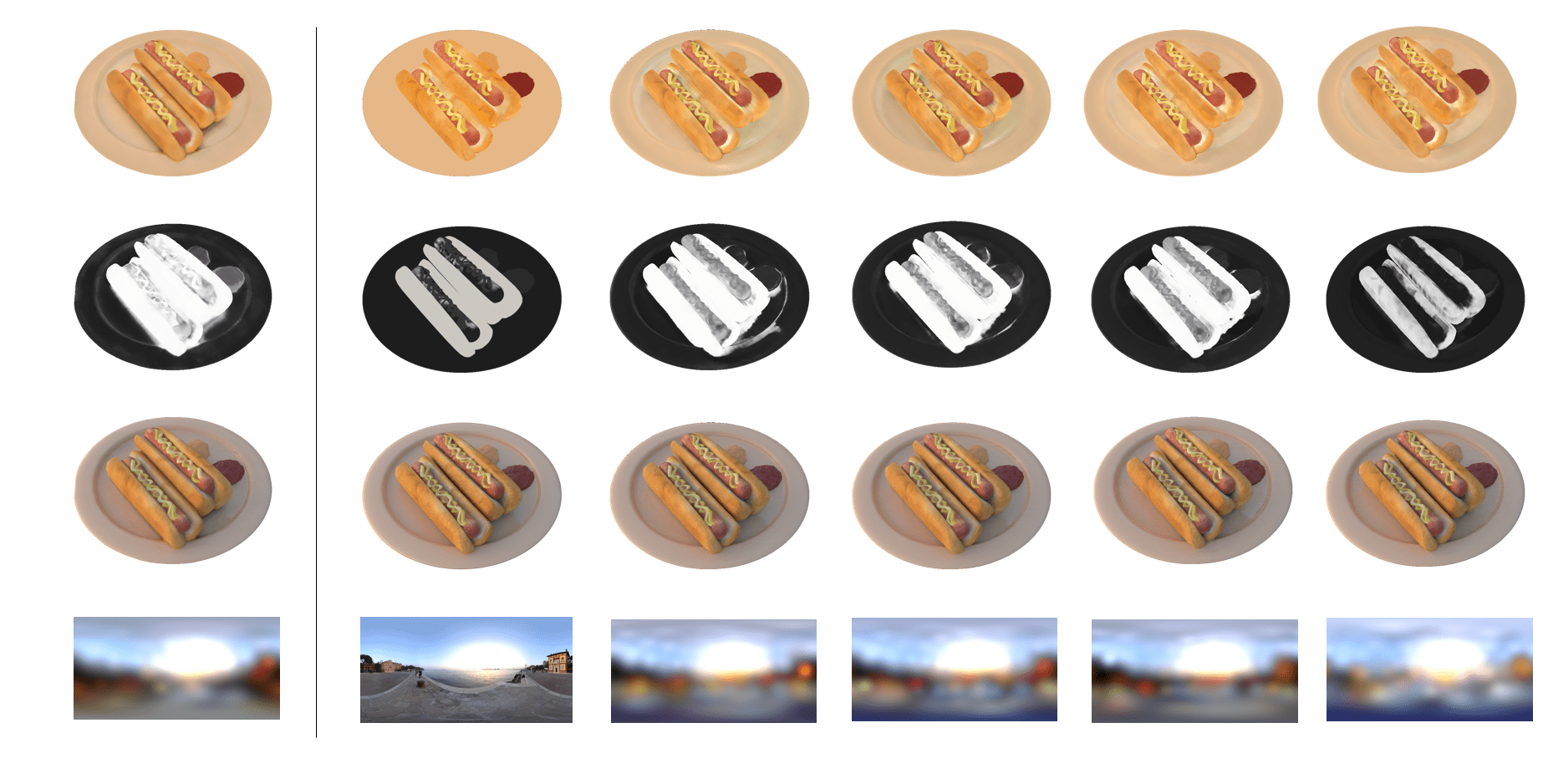}
    \put(9,1){\fontsize{7pt}{0}\selectfont\color{black}{IndiSG}}
    \put(29,1){\fontsize{7pt}{0}\selectfont\color{black}{GT}}
    \put(38.1,1){\fontsize{7pt}{0}\selectfont\color{black}{\textit{w/o} SAI \textit{w/o} VI \textit{w/o} SI}}
    \put(57,1){\fontsize{7pt}{0}\selectfont\color{black}{\textit{w/o} SAI \textit{w/o} VI}}
    \put(74.5,1){\fontsize{7pt}{0}\selectfont\color{black}{\textit{w/o} SAI}}
    \put(91,1){\fontsize{7pt}{0}\selectfont\color{black}{Ours}}
    
    \put(0.5,5.2){\begin{turn}{90}\fontsize{8pt}{0}\selectfont\color{black}{Light}\end{turn}}
    \put(0.5,15.7){\begin{turn}{90}\fontsize{8pt}{0}\selectfont\color{black}{Rendering}\end{turn}}
    \put(0.5,28){\begin{turn}{90}\fontsize{8pt}{0}\selectfont\color{black}{Roughness}\end{turn}}
    \put(0.5,42){\begin{turn}{90}\fontsize{8pt}{0}\selectfont\color{black}{Albedo}\end{turn}}
    \end{overpic}
\caption{Ablation study of material and illumination reconstruction. We show that introducing specular albedo also makes the sausage appear smoother and closer to its true color roughness, represented by black. In terms of lighting, when not using specular albedo, the lighting reconstruction achieves the best result, indicating a clearer reconstruction of ambient illumination.}
\vspace{0.7cm}
\label{fig:ablation_about_hotdog}
\end{figure*}

In \figref{fig:ablation_about_hotdog}, we show that introducing specular albedo also makes the sausage appear smoother and closer to its true color roughness, represented by black. In terms of lighting, when not using specular albedo, the lighting reconstruction achieves the best result, indicating a clearer reconstruction of ambient illumination.
In summary, our ablation study highlights the importance of taking into account various factors when reconstructing materials and illumination from images. By evaluating the performance of different modules, we can better understand their role in improving the reconstruction quality.

\begin{figure*}[h]
\centering
    \begin{overpic}[width=\linewidth]{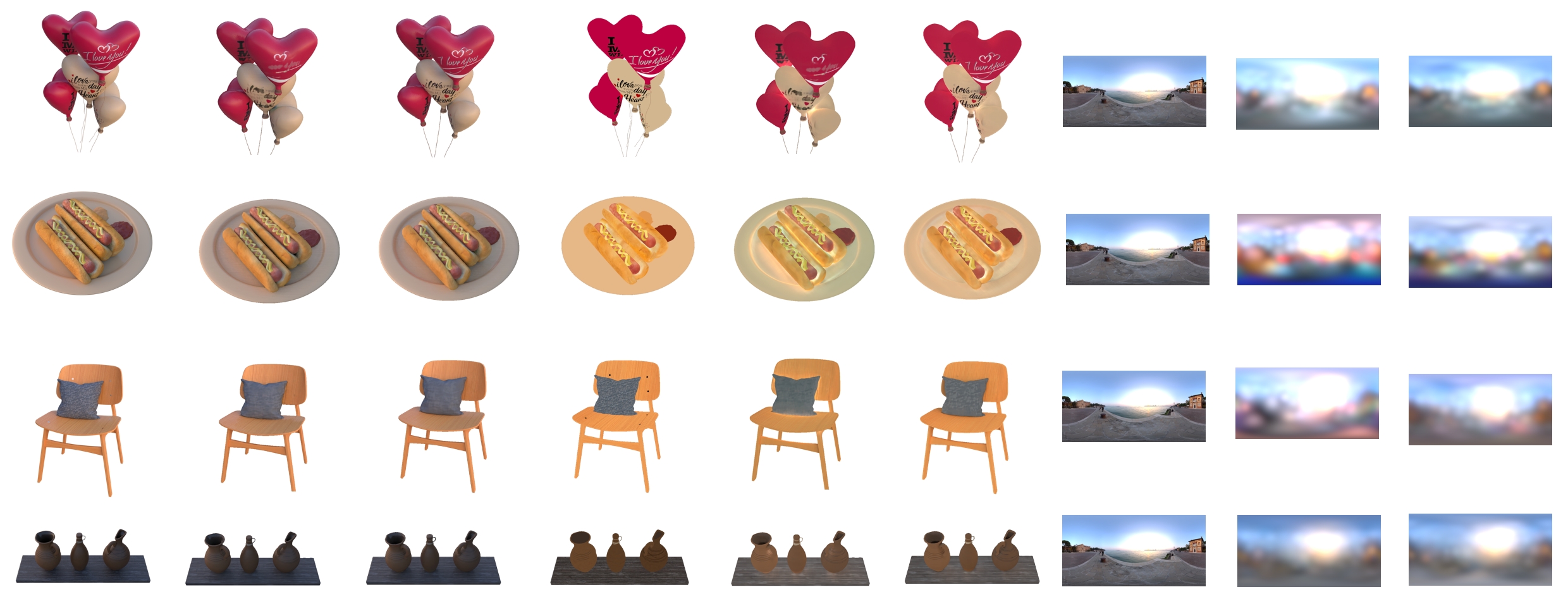}
    
    \put(2,-1){\fontsize{8pt}{0}\selectfont\color{black}{reference}}
    \put(3,-3){\fontsize{8pt}{0}\selectfont\color{black}{image}}
    
    \put(12,-1){\fontsize{8pt}{0}\selectfont\color{black}{w/o indirect}}
    \put(14.5,-3){\fontsize{8pt}{0}\selectfont\color{black}{light}}
    
    \put(26,-1){\fontsize{8pt}{0}\selectfont\color{black}{Ours}}

    \put(36,-1){\fontsize{8pt}{0}\selectfont\color{black}{reference}}
    \put(37,-3){\fontsize{8pt}{0}\selectfont\color{black}{albedo}}
    
    \put(47,-1){\fontsize{8pt}{0}\selectfont\color{black}{w/o indirect}}
    \put(50,-3){\fontsize{8pt}{0}\selectfont\color{black}{light}}
    
    \put(60,-1){\fontsize{8pt}{0}\selectfont\color{black}{Ours}}
    
    \put(69,-1){\fontsize{8pt}{0}\selectfont\color{black}{reference}}
    \put(70.5,-3){\fontsize{8pt}{0}\selectfont\color{black}{light}}
    
    \put(80,-1){\fontsize{8pt}{0}\selectfont\color{black}{w/o indirect}}
    \put(83,-3){\fontsize{8pt}{0}\selectfont\color{black}{light}}

    \put(93, -1){\fontsize{8pt}{0}\selectfont\color{black}{Ours}}

    \end{overpic}
    \vspace{0.3cm}
\caption{Ablation study of indirect light. }
\label{fig:ablation_indiLgt_fig}
\end{figure*}

In stage 3, if we do not consider indirect illumination during the training process, the predicted results for rendering, material, and lighting will all experience a decline. The qualitative results are shown in \figref{fig:ablation_indiLgt_fig}.

\subsection{Additional experiments}
\label{ap:additional_experiments}

To have a fair comparison with Geo-NeuS on the DTU dataset, we incorporate the components of Geo-NeuS based on the additional data (the point clouds from SfM and image pairs) used in Geo-NeuS into our method. As shown in \tabref{tab:chamfer_distance_extra}, our approach can further enhance the surface reconstruction quality on datasets where highlights are less pronounced.

\begin{table}[h]
\caption{Quantitative results in terms of Chamfer distance on DTU~\citep{DTU}.}
\centering
\resizebox{\linewidth}{!}{
\begin{tabular}{lcccc}
\toprule
& DTU 63 & DTU 97 & DTU 110 & \textbf{Mean} \\
\midrule
Geo-NeuS~\cite{geo-neus}  & 0.96 & 0.91 & 0.70  & 0.86 \\
\modelname~(ours)         & 0.99  & 1.15 & 0.89  & 1.01 \\
\modelname~(ours w/ Geo)  & \textbf{0.95} & \textbf{0.89} & \textbf{0.69} & \textbf{0.84} \\
\bottomrule
\end{tabular}
}
\label{tab:chamfer_distance_extra}
\end{table}

\begin{figure*}[t]
\centering
    \begin{overpic}[width=\linewidth]{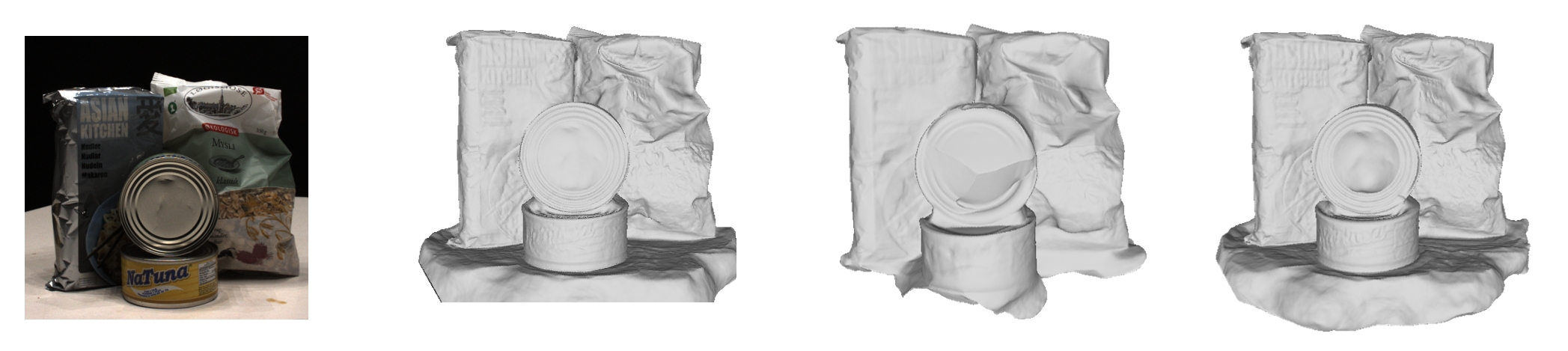}

    \put(6,-2){\fontsize{9pt}{0}\selectfont\color{black}{reference}}
    \put(7.5,-4.5){\fontsize{9pt}{0}\selectfont\color{black}{image}}
    
    \put(35,-2){\fontsize{9pt}{0}\selectfont\color{black}{Ours}}
    \put(35,-4.5){\fontsize{9pt}{0}\selectfont\color{black}{1.15}}
    
    \put(53,-2){\fontsize{9pt}{0}\selectfont\color{black}{Ours with Ref-NeRF}}
    \put(61,-4.5){\fontsize{9pt}{0}\selectfont\color{black}{2.17}}
    
    \put(79,-2){\fontsize{9pt}{0}\selectfont\color{black}{Ours with S$^3$-NeRF}}
    \put(86.5,-4.5){\fontsize{9pt}{0}\selectfont\color{black}{1.23}}
    
    \end{overpic}
    \vspace{0.5cm}
\caption{Comparison with Ref-NeRF and S$^3$-NeRF. }
\label{fig:compare_refnerf_s3nerf}
\end{figure*}

We conduct another experiment to compare our modeling approach with Ref-NeRF and S$^3$-NeRF~\citep{s3nerf}. The experimental quantitative and qualitative results are shown in \figref{fig:compare_refnerf_s3nerf}. Ref-NeRF utilizes volume rendering colors for diffuse and specular components. If we directly combine SDF and the architecture of Ref-NeRF, it is challenging to eliminate the influence of highlights.
Furthermore, if we applied the construction method of S$^3$-NeRF, which involves integrating surface rendering colors into volume rendering, to modify our model structure, we found that this modeling approach cannot address the issue of geometric concavity caused by highlights.

\begin{figure*}[t]
\centering
    \begin{overpic}[width=\linewidth]{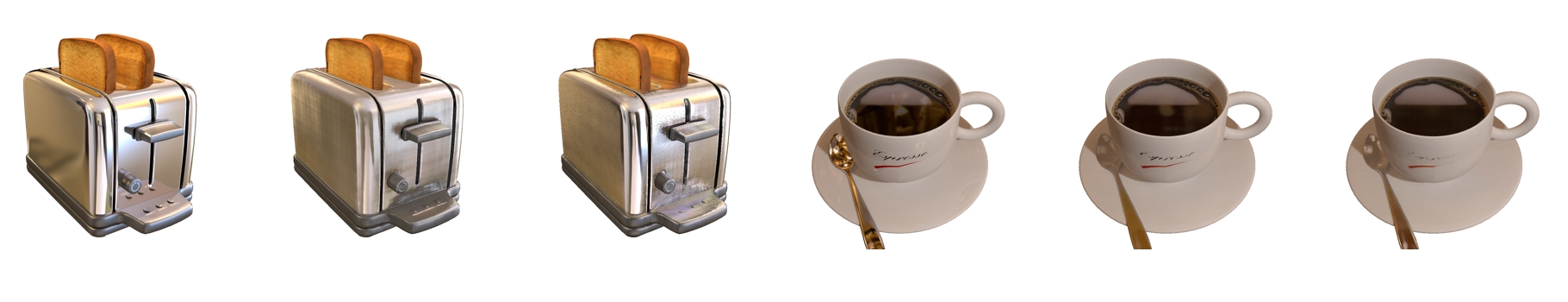}
    
    \put(4,0){\fontsize{8pt}{0}\selectfont\color{black}{GT image}}
    
    \put(22.5,0){\fontsize{8pt}{0}\selectfont\color{black}{f=0.02}}
    \put(22.7,-2){\fontsize{8pt}{0}\selectfont\color{black}{22.33}}
    
    \put(40.2,0){\fontsize{8pt}{0}\selectfont\color{black}{f=0.75}}
    \put(40.4,-2){\fontsize{8pt}{0}\selectfont\color{black}{24.31}}

    \put(54.8,0){\fontsize{8pt}{0}\selectfont\color{black}{GT image}}
    
    \put(72.5,0){\fontsize{8pt}{0}\selectfont\color{black}{f=0.02}}
    \put(72.7,-2){\fontsize{8pt}{0}\selectfont\color{black}{30.43}}
    
    \put(90,0){\fontsize{8pt}{0}\selectfont\color{black}{f=0.75}}
    \put(90.2,-2){\fontsize{8pt}{0}\selectfont\color{black}{30.18}}
    
    \end{overpic}
    \vspace{0cm}
\caption{Adjusting Fresnel value to model chrome-like appearance.}
\label{fig:fresnel}
\end{figure*}

For chrome-like materials, We increase the Fresnel value to 0.75 in the rendering formula of stage 3 to test the impact of this operation. We show the results and their PSNR value in \figref{fig:fresnel}, we observed that increasing the Fresnel value indeed leads to the better reconstruction of objects with chrome-like materials. For the Toaster model, we observed a significant improvement in PSNR with an increased Fresnel value. However, we also noticed that solely increasing the Fresnel value can result in the degradation of texture details. For instance, in the Coffee model, although the highlights on the spoon are better reconstructed, the text on the cup deteriorates. One of our future directions is to address this issue more effectively.

\section{Limitations}
\label{ap:}

\begin{figure*}[t]
\vspace{-0.1cm}   
\centering
    \begin{overpic}[width=\linewidth]{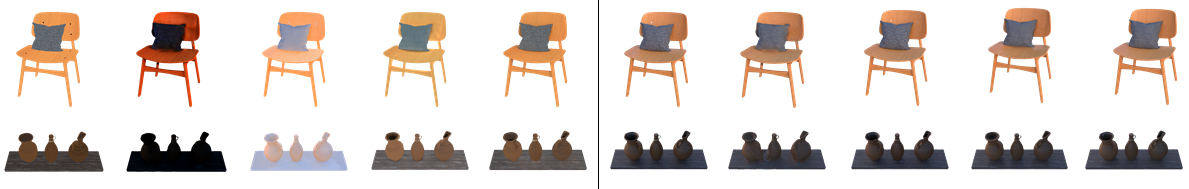}
    \put(1.5,-1){\fontsize{8pt}{0}\selectfont\color{black}{GT albedo}}
    \put(12.5,-1){\fontsize{8pt}{0}\selectfont\color{black}{PhySG}}
    \put(21.5,-1){\fontsize{8pt}{0}\selectfont\color{black}{NVDiffrec}}
    \put(33,-1){\fontsize{8pt}{0}\selectfont\color{black}{IndiSG}}
    \put(43.8,-1){\fontsize{8pt}{0}\selectfont\color{black}{Ours}}
    \put(52.5,-1){\fontsize{8pt}{0}\selectfont\color{black}{GT image}}
    \put(63.2,-1){\fontsize{8pt}{0}\selectfont\color{black}{PhySG}}
    \put(72.3,-1){\fontsize{8pt}{0}\selectfont\color{black}{NVDiffrec}} 
    \put(83.5,-1){\fontsize{8pt}{0}\selectfont\color{black}{IndiSG}} 
    \put(93.7,-1){\fontsize{8pt}{0}\selectfont\color{black}{Ours}}

    \put(-1.3,1.5){\begin{turn}{90}\fontsize{8pt}{0}\selectfont\color{black}{Jugs}\end{turn}}
    \put(-1.3,10){\begin{turn}{90}\fontsize{8pt}{0}\selectfont\color{black}{Chair}\end{turn}}
    \end{overpic}
    \vspace{-0.3cm}
\caption{Qualitative results for other scenes (chair and jugs) on IndiSG dataset in terms of albedo reconstruction (left) and novel view synthesis quality (right).}  
\label{fig:decomp_indi}
\end{figure*}

In certain scenarios, our method still faces difficulties.
For mesh reconstruction, 
despite improvements on the glossy parts in the DTU 97 tin model, the overall Chamfer distance does not significantly decrease due to the small proportion of glossy parts. However, for scenes with large areas of glossy parts, such as the flower pot model, our improvements are more pronounced and surpass Geo-NeuS.
As seen in Appx~\figref{fig:decomp_indi}, the reconstructed albedo of the chair still lacks some detail.
The nails on the chair and the textures on the pillow are not accurately captured in the reconstructed geometry. A future research direction is how to effectively decompose materials for fine structures, such as nails on the backrest of a chair.


\end{document}

%% file: tables/chamfer_distance.tex
\begin{tabular}{lcccccccccc}
\toprule
& Pot & Funnel & Snowman &Jug & \textbf{Mean} & DTU 63 & DTU 97 & DTU 110 & DTU 122 & \textbf{Mean}\\
\midrule
Geo-NeuS (PC)~\cite{geo-neus}  & \cellcolor{secondorange}{1.88} & \cellcolor{secondorange}{2.03} & \cellcolor{secondorange}{1.64} & \cellcolor{secondorange}{1.68}  & \cellcolor{secondorange}{1.81} & \cellcolor{firstred}{0.96} & \cellcolor{firstred}{0.91} & \cellcolor{firstred}{0.70}  & \cellcolor{firstred}{0.37} & \cellcolor{firstred}{0.73} \\
\midrule
PhySG~\cite{Physg}  & 14.40 & 7.39 & 1.55 & 7.59  & 7.73 & 4.16 & 4.99 & 3.57 & 1.42  & 3.53 \\
IndiSG~\cite{Indirect}  & 5.62 & 4.05 & 1.74 & 2.35  & 3.44 
 & 1.15 & 2.07 & 2.60 & 0.61  & 1.61\\
NeuS~\cite{NeuS}  & 2.09 & 3.93 & 1.40 & 1.81  & 2.31 & 1.01 & 1.21 & 1.14  & 0.54 & 0.98 \\
NeRO~\cite{nero}  & 6.03 & 2.63 & 1.71 & 4.23  & 3.65 & 1.32 & 1.47 & 1.14  & 0.57 & 1.12 \\

\modelname~(ours) & \cellcolor{firstred}{1.54} & \cellcolor{firstred}{1.95} & \cellcolor{firstred}{1.31} & \cellcolor{firstred}{1.40}  & \cellcolor{firstred}{1.55} & \cellcolor{secondorange}{0.99}  & \cellcolor{secondorange}{1.15} & \cellcolor{secondorange}{0.89} & \cellcolor{secondorange}{0.46} & \cellcolor{secondorange}{0.87} \\
\bottomrule
\end{tabular}

%% file: tables/nerfsyn_var.tex
\begin{tabular}{lccccccccccccccc}
\toprule
& \multicolumn{3}{c}{Baloons} & \multicolumn{3}{c}{Hotdog} & \multicolumn{3}{c}{Chair} & \multicolumn{3}{c}{Jugs} & \multicolumn{3}{c}{\textbf{Mean}} \\
& albedo & illumination & rendering & albedo & illumination & rendering & albedo & illumination & rendering & albedo & illumination & rendering & albedo & illumination & rendering \\
\midrule
PhySG  & 15.91 & 13.89  & 27.83  & 13.95 & 11.69  & 25.13  & 14.86 & 12.26  & 28.32 & 16.84 & 10.92 & 28.20 & 15.39 & 12.19 & 27.37\\
NVDiffrec  & \bf{26.88} & 14.63  & 29.90 & 13.60 & 22.43  & 33.68  & 21.12 & 15.56  & 29.16 & 11.20 & 10.47 & 25.30 & 20.41 & 13.56 & 29.51 \\
IndiSG  & 21.95 & 25.24  & 24.40 & 26.43 & 21.87  & 31.77  & 24.71 & \bf{22.17}  & 24.98 & 21.44 & 20.59 & 24.91 & 23.63 & 22.47 & 26.51 \\
Ours \textit{w/o} IndiLgt & 23.13 & 18.24  & 29.45  & 25.62 & 17.97 & 35.97 & 25.22 & 18.04 & 34.31 & 22.87 & 21.84 & 26.30 & 24.21 & 19.02 & 31.51 \\
Ours \textit{w/o} SAI  & 24.09 & \bf{25.97}  & 28.82 & 30.58 & \bf{23.50}  & 36.05  & 25.23 & 22.13  & 32.64 & 19.64 & 20.40 & 33.56 & 24.89 & \bf{23.00} & 32.77 \\
Ours & 25.79 & 21.79  & \bf{33.89} & \bf{30.72} & 20.23 & \bf{36.71}  & \bf{26.33} & 20.97  & \bf{34.58} & \bf{22.94} & \bf{21.84} & \bf{36.48} & \bf{26.28} & 21.21 & \bf{35.41} \\
\bottomrule
\end{tabular}

%% file: tables/ablation_hotdog.tex
\begin{tabular}{lccccc}
\toprule
Method &  Alb & Rough & Rend & Illu \\
\midrule
IndiSG~\cite{Indirect} &	                       26.44 & 15.97 & 31.78 & 21.88 \\
\midrule
Ours \textit{w/o} SAI \textit{w/o} VI \textit{w/o} SI    &	29.31 & 16.98 &	35.48  & 23.48 \\
Ours \textit{w/o} SAI \textit{w/o} VI     &	29.64 & 17.86 & 36.36 &	23.41  \\
Ours \textit{w/o} SAI  &	               30.58 & 18.83 & 36.05 &	\textbf{23.50} \\
Ours &	                       \textbf{30.76} & \textbf{23.10} & \textbf{36.71} & 20.24 \\
\bottomrule
\end{tabular}